\documentclass[twoside]{article}

\usepackage[preprint]{aistats2027}
\usepackage{amsmath,amssymb,mathtools}
\usepackage{amsthm}
\usepackage{booktabs}
\usepackage{placeins}
\usepackage{graphicx}
\usepackage{xcolor}
\usepackage[colorlinks=true,linkcolor=blue!60!black,citecolor=blue!60!black,urlcolor=blue!60!black]{hyperref}

\usepackage[round]{natbib}

\newtheorem{theorem}{Theorem}
\newtheorem{proposition}[theorem]{Proposition}
\newtheorem{corollary}[theorem]{Corollary}
\newtheorem{lemma}[theorem]{Lemma}

\theoremstyle{definition}

\theoremstyle{remark}
\newtheorem{remark}[theorem]{Remark}
\theoremstyle{plain}

\begin{document}

%
\runningtitle{Patch Attention Wisely}

%

\runningauthor{Junyu Chen, Likai Chen, Weining Wang}

\twocolumn[

\aistatstitle{Patch Attention Wisely: Understanding Patch Length in Attention-Based Forecasting}

\aistatsauthor{Junyu Chen\textsuperscript{*} \And Likai Chen \And Weining Wang}

\aistatsaddress{Department of Economics,\\ Econometrics and Finance\\University of Groningen \And Department of Statistics\\ and Data Science\\ Washington University in St.~Louis \And School of Economics\\ University of Bristol} ]

\begingroup
    \renewcommand{\thefootnote}{\fnsymbol{footnote}}
    \footnotetext[1]{\raggedright Correspondence to: \href{mailto:junyu.chen@rug.nl}{\texttt{junyu.chen@rug.nl}}.}
\endgroup

\begin{abstract}
Attention-based time series forecasters commonly group consecutive observations into patches. Patch length is known to have a substantial impact on forecasting performance, yet its selection remains largely empirical. We develop a theoretical framework that explains why patch length matters for forecasting accuracy and provides a principled basis for its selection. For linear state-space models, we derive an exact factorization of the forecasting coefficient matrix that separates forecast dynamics from filtering dynamics. We show that dominance of a single forecast component at long horizons does not generally imply a rank-one representation. This explains why longer horizons alone need not reduce representational complexity. We combine the factorization with estimation bounds to develop a patch-selection criterion that balances representation and estimation errors under temporal dependence. Our analysis establishes conditions under which balanced patching is justified and clarifies how predictive structure, head count, training sample size, and long-run dependence enter the selection criterion.
\end{abstract}

\section{INTRODUCTION}
\label{sec:introduction}
Patching has become a basic design choice in attention-based time-series forecasting. A history of length $L$ can be presented as $L$ point tokens or as $N=L/p$ non-overlapping patches of length $p$. Increasing $p$ shortens the token sequence and can make long contexts computationally feasible. Within one attention head, a shared linear value map applies the same $p$ within-patch weights to every patch, and the query--key scores only rescale this pattern from patch to patch. Patch length is therefore not only a computational choice: it also determines which forecasts a fixed number of heads can express and how many coefficients must be learned from data. With limited data or long histories, an unsuitable patch length can leave too many coefficients to estimate or too little flexibility to fit the target, even when it reduces attention cost.

Prior work has established that patch tokens can be both accurate and efficient. PatchTST \citep{nie2023patchtst} showed that patching preserves local temporal structure, reduces the number of tokens, and permits longer look-back windows, and patch tokenization is now used in large pretrained forecasters such as TimesFM \citep{das2024timesfm}. It has also shown that a single temporal resolution need not suffice: MTST \citep{zhang2024mtst} uses parallel patch scales, Pathformer \citep{chen2024pathformer} combines scales through adaptive pathways, and TimeSqueeze \citep{ankireddy2026timesqueeze} moves patch boundaries according to local signal complexity. Complementary studies compare point-wise, patch-wise, and variate-wise tokens \citep{chen2025closer}, and a controlled analysis finds that adaptive patching should be compared against carefully tuned uniform patches \citep{zucchi2026adaptive}. These works explain how to build effective tokenizers and when they perform well empirically.

Most existing work on patching is algorithmic and empirical: it proposes tokenization schemes and compares their accuracy and computational cost, but it provides no statistical theory for choosing a fixed patch length for a given forecasting model. Prediction errors alone cannot isolate the effect of patch length, because changing $p$ also changes the number of tokens, how parameters are shared across time, the number of coefficients to estimate, the optimization problem, and the computational cost. Multiscale and adaptive architectures face the same ambiguity. Moreover, non-overlapping patching does not discard any observation but only regroups the observations into patches, so any loss of forecasting power must come from the model applied after this regrouping. A statistical theory of patch selection must therefore first characterize how patch length changes that model and then determine when one patch length should be preferred.

We provide such a characterization for a multi-head linear attention model. Each head applies one set of within-patch weights, and its query--key scores determine how strongly these weights are used in each patch. Summed over heads, these contributions form an $N\times p$ \emph{coefficient matrix} that contains the weight on every observation in the history, and with $m$ heads this matrix has rank at most $m$. When the attention weights do not depend on the input and every patch position has its own learnable key, the model can represent exactly the coefficient matrices of rank at most $m$. Choosing $p$ therefore reshapes the same length-$L$ linear forecast into a different matrix, which can change both the rank needed to represent it and the difficulty of estimating it. This reduction to reduced-rank regression isolates the effect of patch length without claiming to describe every component of a full Transformer.

A central difficulty is that a long horizon does not make the forecast simple to represent. In a partially observed dynamical system, a long-horizon forecast may be dominated by one slowly evolving state component, and it is tempting to conclude that one weighting pattern then suffices. We show that this is not true, because the forecaster must first recover that slow component from observations contaminated by faster latent dynamics and measurement noise. Using a linear state-space model, we factor the optimal forecast coefficients into horizon-dependent weights and horizon-independent filtering patterns. The factorization shows that the component dominating the distant future can still require several filtering patterns to recover. It also identifies \emph{patch aliasing}, in which a patch length makes two different filtering patterns produce identical weights across patches. Consequently, the best low-rank approximation can keep a positive relative error as the horizon grows.

We relate the infinite-past coefficients to the best forecast from a finite window and bound the error of rank-constrained least squares fitted to overlapping windows of one dependent series. The bound separates approximation error from an estimation term governed by the dimension of the rank-constrained coefficient class. When the rank constraint is active, this dimension is $m(L/p+p-m)$, which is smallest when the patch length and the number of patches are equal. This suggests patch lengths near $\sqrt L$ when approximation errors are comparable, although with enough data a less balanced reshape can be preferable if it represents the target more accurately.

In summary, we contribute \textbf{(i)} an exact identity linking head count and patch length to the coefficient class of the model, \textbf{(ii)} a state-space factorization and a long-horizon floor on the relative approximation error, which separate the components that drive the forecast from the filtering patterns needed to recover them, \textbf{(iii)} finite-window and dependent-data bounds that expose the approximation--estimation tradeoff in patch selection, and \textbf{(iv)} a mixture setting in which input-dependent attention weights can improve the forecast because the observed history reveals which system generated it. Simulations show that intermediate patch lengths near $\sqrt L$ help most when training data are scarce, and that their advantage narrows as the sample grows.

\paragraph{Related work.}
Our question differs from showing that attention can implement a Kalman filter or learn an autoregressive rule in context \citep{goel2024transformer,sander2024autoregressive,lu2025var}: we study how a specified forecasting model distributes coefficients over a patched history. The non-overlapping coefficient reshape is related to Page-matrix methods, which use low-rank structure for time-series estimation and prediction \citep{damen1982page,agarwal2018model}, and contrasts with the overlapping Hankel matrices common in realization theory \citep{glover1984hankel}. Spectral learning of dynamical systems provides complementary low-dimensional descriptions of long-memory predictors \citep{hazan2017spectral,rashidinejad2020slip}.  We combine these structural ideas with reduced-rank regression \citep{izenman1975rrr,velu1987rrr} and dependent time-series estimation \citep{basu2015regularized}. The new link is that the rank constraint is induced directly by attention heads and changes shape with the patch length.

\section{PATCH ATTENTION AS REDUCED-RANK REGRESSION}
\label{sec:setup}

In this section, we connect attention to reduced-rank regression and separate approximation error from estimation error.

Let $(X_t)$ be a centered stationary scalar process with finite second moments. We forecast $Y_t=X_{t+H}$, $H\ge1$, from the window $x_{t,L}=(X_t,\ldots,X_{t-L+1})^\top$. Let $p$ denote the patch length, assume that $p$ divides $L$, and write $N=L/p$ for the number of non-overlapping patch tokens. Define
\begin{equation}
    Z_{t,p}[j,s]=X_{t-(jp+s)}, 
    \quad 0\le j<N,\quad 0\le s<p.
    \label{eq:patch}
\end{equation}
Thus $Z_{t,p}\in\mathbb{R}^{N\times p}$, with rows ordered from recent to older patches and observations within each row ordered in reverse chronological order. Patching rearranges the window without discarding observations.

To describe the attention mechanism, let $\mathbf T\in\mathbb{R}^{N\times d}$ contain token embeddings incorporating patch content and position. Head $a$ forms the arrays
\begin{equation}
    \mathbf Q_a=\mathbf T\widetilde W_{Q,a}, \quad \mathbf K_a=\mathbf T\widetilde W_{K,a}, \quad \mathbf V_a=\mathbf T\widetilde W_{V,a},
\end{equation}
where $\mathbf Q_a,\mathbf K_a\in\mathbb{R}^{N\times d_k}$ and $\mathbf V_a\in\mathbb{R}^{N\times d_v}$. Standard dot-product attention normalizes query--key similarities through a softmax. To obtain a tractable model, we replace this normalization by unnormalized scores with a fixed factor $1/N$. The resulting all-token attention operation is
\begin{equation}
    \mathbf O_a = \frac1N(\mathbf Q_a\mathbf K_a^\top)\mathbf V_a = \frac1N\mathbf Q_a(\mathbf K_a^\top\mathbf V_a).
\end{equation}
The scores may be signed and need not sum to one, and the factor $1/N$ depends only on the token count.

For forecasting, we specialize this operation to one forecast query per head, and we compute each value by applying the same linear map to the raw patch. Write $x_j\in\mathbb{R}^p$ for the transpose of row $j$ of an input $Z$. A forecast-token representation $c_H(Z)\in\mathbb{R}^d$, constructed from the observed window and horizon, supplies the query $q_a(Z)=\widetilde W_{Q,a}^\top c_H(Z)$. Let $k_{a,j}(Z)$ be the column vector corresponding to row $j$ of $\mathbf K_a$. We take $v_{a,j}(Z)=W_{V,a}x_j$, where the same matrix $W_{V,a}\in\mathbb{R}^{d_v\times p}$ is used for every patch. A linear patch embedding that is common to all patches can be absorbed into $W_{V,a}$. The analyzed forecast consists solely of these attention outputs combined through bias-free linear output weights $w_a\in\mathbb{R}^{d_v}$. With $\alpha_{a,j}(Z)=N^{-1}k_{a,j}(Z)^\top q_a(Z)$, it is
\begin{equation}
    \widehat f(Z) = \sum_{a=1}^m\sum_{j=0}^{N-1} \alpha_{a,j}(Z)\,w_a^\top W_{V,a}x_j.
    \label{eq:model}
\end{equation}
Because every patch passes through the same value and output maps, the coefficients on the raw observations have the following structure.

\begin{proposition}[Coefficient matrix]
\label{prop:routing}
Let $b_a=W_{V,a}^\top w_a\in\mathbb{R}^p$ and $\alpha_a(Z)=(\alpha_{a,0}(Z),\ldots,\alpha_{a,N-1}(Z))^\top$. Then
\begin{equation}
    \widehat f(Z)=\langle B(Z),Z\rangle_{\mathrm{F}}, \qquad B(Z)=\sum_{a=1}^m\alpha_a(Z)b_a^\top,
\end{equation}
and $\operatorname{rank} B(Z)\le\min\{m,N,p\}$ for every input $Z$.
\end{proposition}

Each head supplies one within-patch weighting pattern $b_a$, whose contribution across patches is determined by $\alpha_a(Z)$. Unlike the $N\times N$ query--key score matrix, $B(Z)$ acts directly on the observations. Its rank bound holds even for input-dependent attention weights, although $B(Z)$ then varies with $Z$ and the forecast is generally nonlinear.

For tractability, we restrict the model to fixed attention weights: each head uses a constant forecast query and keys that depend only on the patch position, so the weights do not depend on the input and $B(Z)=B$. This restriction suits linear state-space forecasting, whose target is the best linear forecast from the window, and allows representation and estimation errors to be separated explicitly. It is an additional assumption, not a consequence of Proposition~\ref{prop:routing}.

If every patch position has its own learnable key and the value and output maps are unrestricted, every $N\times p$ matrix of rank at most $m$ can be represented with $m$ heads (Appendix~\ref{app:setup-proofs}). Thus the predictor class induced by Equation~\eqref{eq:model} with fixed attention weights is exactly
\begin{equation}
    \mathcal{M}_m(p) = \{B\in\mathbb{R}^{N\times p}:\operatorname{rank}(B)\le m\}.
\end{equation}
Fitting this model by least squares is therefore reduced-rank regression on the reshaped window.

Let $b^\circ\in\mathbb{R}^L$ denote the coefficient vector of the best linear forecast of $Y_t=X_{t+H}$ from $x_{t,L}$, and let $R_0$ denote its mean squared prediction error. Assume $\Gamma_L=\operatorname{Cov}(x_{t,L})\succ0$. For each divisor $p$ of $L$, let $B_p^\circ\in\mathbb{R}^{N\times p}$ be the corresponding reshape:
\begin{equation}
    B_p^\circ[j,s]=b^\circ_{jp+s}, \qquad 0\le j<N, \quad 0\le s<p.
\end{equation}

\begin{lemma}[Reshaping invariance]
\label{lem:reshape}
With column-stacking vectorization, there is a permutation matrix $\Pi_p$ such that $\operatorname{vec}(Z_{t,p})=\Pi_p x_{t,L}$. Consequently,
\begin{equation}
    \Gamma_{L,p} := \operatorname{Cov}\{\operatorname{vec}(Z_{t,p})\} = \Pi_p\Gamma_L\Pi_p^\top
\end{equation}
has the same eigenvalues as $\Gamma_L$ for every $p$. For every $B\in\mathbb{R}^{N\times p}$,
\begin{equation}
    \mathbb{E}(Y_t-\langle B,Z_{t,p}\rangle_{\mathrm{F}})^2 = R_0+\|B-B^{\circ}_{p}\|_\Gamma^2,
    \label{eq:risk}
\end{equation}
where $\|D\|_\Gamma^2=\operatorname{vec}(D)^\top\Gamma_{L,p}\operatorname{vec}(D)$. In particular, the minimum risk over unrestricted linear forecasts is $R_0$ for every patch length.
\end{lemma}

Patching does not change the best linear forecast, but it changes the shape of the coefficient matrix that $m$ heads must approximate. Define the target rank and covariance-weighted approximation error by
\begin{equation}
    r_p=\operatorname{rank}B^{\circ}_{p}, \quad A_m(p)=\min_{B\in\mathcal{M}_m(p)}\|B-B^{\circ}_{p}\|_\Gamma^2.
\end{equation}
The approximation error vanishes if and only if $m\ge r_p$. Otherwise, its size depends on both the reshaped coefficients and the covariance of the observations. Patch selection balances this error against the difficulty of estimating a matrix in $\mathcal{M}_m(p)$ (Section~\ref{sec:selection}).

\section{FORECAST HORIZON AND APPROXIMATION LOSS UNDER A STATE-SPACE MODEL}
\label{sec:model}

Section~\ref{sec:setup} characterizes forecasting loss through the target coefficients and their approximation by attention heads, but leaves the role of horizon implicit. To characterize this loss further, we use a state-space model as a tractable benchmark. Forecasting then involves estimating the current latent state from noisy observations and propagating it to the forecast horizon. This separation makes the horizon dependence of the coefficients explicit, allowing us to study their rank and approximation loss under patching. In particular, recovering a single dominant state component at long horizons can still require several temporal filtering patterns, so its patch representation need not have rank one.

\subsection{The State-Space Model and Its Forecast Coefficients}

Let
\begin{equation}
 S_{t+1}=AS_t+\eta_{t+1},\qquad X_t=CS_t+\xi_t,
 \label{eq:ssm}
\end{equation}
where $S_t\in\mathbb{R}^r$, $A\in\mathbb{R}^{r\times r}$, and $C\in\mathbb{R}^{1\times r}$. The centered noises are mutually independent and independent across time, with finite second moments, process covariance $Q\succeq0$, and measurement variance $R>0$. Assume $\rho_A<1$, where $\rho_M$ denotes the spectral radius of a matrix $M$, and stationary initialization.

To expose the filtering structure, first consider prediction from the entire observed past. Let $K\in\mathbb{R}^{r\times1}$ be the steady-state Kalman predictor gain and set $F=A-KC$. The best linear predictor $m_t$ of $S_t$ based on $(X_s:s<t)$ satisfies
\begin{equation}
 m_{t+1}=Fm_t+KX_t,\qquad \rho_F<1.
 \label{eq:observer}
\end{equation}
Consequently, the best linear forecast from the infinite past is
\begin{align}
\widehat X_{t+H\mid t}
 &=\sum_{\ell\ge0}\beta_H(\ell)X_{t-\ell}, \nonumber\\
\beta_H(\ell)&=CA^{H-1}F^\ell K. \label{eq:kernel}
\end{align}
We now split the coefficients in \eqref{eq:kernel} into filtering modes. A mode is a dynamical component associated with an eigenvalue and its eigendirection. Filtering modes are the modes of $F$, whose powers describe how past observations reveal the current state, and forecast modes are the modes of $A$, which carries that estimate to the forecast horizon. This separates the effect of the horizon, which changes their weights, from the effect of patch length, which changes how they are arranged. These two roles need not have the same effective dimension.

Keep the latest $L$ coefficients and arrange them like the input:
\begin{equation}
 B_{H,p}[j,s]=\beta_H(jp+s).
 \label{eq:patchmatrix}
\end{equation}
The matrix $B_{H,p}$ keeps the first $L$ coefficients of the infinite-past forecast, whereas $B_p^\circ$ is the best forecast that uses only the window. We also write $B^\circ_{H,L,p}=B_p^\circ$ when displaying the horizon and window length explicitly. Theorem~\ref{thm:bridge} in Appendix~\ref{sec:estimation} bounds this difference, so the spectral results for $B_{H,p}$ below also describe $B_p^\circ$.

\subsection{From Filtering Modes to Patch Patterns}
\label{sec:spectra}
We next express the coefficients in \eqref{eq:kernel} through filtering modes and connect these modes to patch patterns. Assume that $A$ and $F$ have distinct eigenvalues and are diagonalizable over $\mathbb C$. Write $A=\sum_{a=1}^r\lambda_aP_a$ and $F=\sum_{k=1}^r\varphi_kQ_k$, where $P_a$ and $Q_k$ are the corresponding spectral projectors. Define
\begin{equation}
 g_{ak}=CP_aQ_kK,\qquad
 d_k(H)=\sum_{a=1}^r g_{ak}\lambda_a^{H-1},
 \label{eq:coupling}
\end{equation}
and the geometric patterns
\begin{align}
u_{N,p}(z)&=(1,z^p,\ldots,z^{(N-1)p})^\top, \nonumber\\
 v_p(z)&=(1,z,\ldots,z^{p-1})^\top.
\end{align}

\begin{theorem}[Filtering-mode factorization]
\label{thm:factor}
For every $H\ge1$,
\begin{equation}
 B_{H,p}=\sum_{k=1}^r d_k(H)
 u_{N,p}(\varphi_k)v_p(\varphi_k)^\top,
 \label{eq:factor}
\end{equation}
so $\operatorname{rank}(B_{H,p})\le\min\{r,N,p\}$. The filtering factors are independent of the forecast horizon.
\end{theorem}
Each filtering mode contributes one across-patch pattern and one within-patch pattern, since $\varphi^{jp+s}=(\varphi^p)^j\varphi^s$. The horizon changes their weights $d_k(H)$, rather than their shapes. Complex terms occur in conjugate pairs and sum to a real matrix. Appendix~\ref{app:spectral-proofs} provides the proof.

\subsection{A Relative Filtering Floor (\texorpdfstring{$H \to \infty$}{H to infinity})}
We now show that a low-rank patch representation can keep a non-vanishing relative approximation error, even when one forecast mode dominates as $H\rightarrow\infty$. To see this, suppose one forecast eigenvalue dominates: $\lambda_1\ne0$ is simple and $|\lambda_1|>\max_{a\ge2}|\lambda_a|$, with $r\ge2$. Set
\begin{align}
q_A&=\max_{a\ge2}|\lambda_a/\lambda_1|<1, \nonumber\\
 B_{\infty,p}&=\sum_{k=1}^r g_{1k}
 u_{N,p}(\varphi_k)v_p(\varphi_k)^\top,
\end{align}
and assume $B_{\infty,p}\ne0$. Let $C_p=\sum_{a=2}^r\sum_{k=1}^r|g_{ak}|\|u_{N,p}(\varphi_k)\|_2\|v_p(\varphi_k)\|_2$. Rescaling by $\lambda_1^{-(H-1)}$ removes the decay of the dominant forecast mode and reveals the limiting coefficient matrix $B_{\infty,p}$. For a rank budget $h$, the following theorem shows that the relative approximation error in operator norm converges to that of this matrix. Thus a longer horizon need not remove the relative error caused by using too few heads.
\begin{theorem}[Long-horizon filtering floor]
\label{thm:floor}
For every $H\ge1$,
\begin{equation}
 \|\lambda_1^{-(H-1)}B_{H,p}-B_{\infty,p}\|_{\mathrm{op}}
 \le C_pq_A^{H-1}.
 \label{eq:oprate}
\end{equation}
Let $R^{\mathrm{op}}_{H,p,h}=\sigma_{h+1}(B_{H,p})/\sigma_1(B_{H,p})$ and $\kappa_{\infty,h}(p)=\sigma_{h+1}(B_{\infty,p})/\sigma_1(B_{\infty,p})$. If $C_pq_A^{H-1}\le\sigma_1(B_{\infty,p})/2$, then
\begin{equation}
 |R^{\mathrm{op}}_{H,p,h}-\kappa_{\infty,h}(p)|
 \le\frac{4C_pq_A^{H-1}}{\sigma_1(B_{\infty,p})}.
 \label{eq:relrate}
\end{equation}
The limit is positive when $h<\operatorname{rank}(B_{\infty,p})$.
\end{theorem}
The limiting rank depends on how many filtering modes contribute to the dominant forecast component. If the patch matrix has enough rows and columns, and these modes remain distinct both across patches and within each patch, their patterns are linearly independent. The following corollary identifies the limiting rank with the number of active filtering modes.
\begin{corollary}[Active filtering modes and patch aliasing]
\label{cor:active-modes}
Let $\mathcal S=\{k:g_{1k}\ne0\}$. If $|\mathcal S|\le\min\{N,p\}$ and the nodes $\{\varphi_k:k\in\mathcal S\}$ and $\{\varphi_k^p:k\in\mathcal S\}$ are each pairwise distinct, then $\operatorname{rank}(B_{\infty,p})=|\mathcal S|$.
\end{corollary}

The active modes depend on the system, while patching determines whether their across-patch patterns remain distinct. For example, $\varphi$ and $-\varphi$ produce identical across-patch patterns when $p$ is even. We call this coincidence \emph{patch aliasing}.

The floor is \emph{relative}. Writing $\mathcal{E}_h^\Gamma(B)=\min_{\operatorname{rank} M\le h}\|B-M\|_\Gamma^2$, for fixed $L,p,h$,
\begin{equation}
 \mathcal{E}_h^\Gamma(B_{H,p})=|\lambda_1|^{2(H-1)}
 \{\mathcal{E}_h^\Gamma(B_{\infty,p})+O(q_A^{H-1})\}.
 \label{eq:absolute}
\end{equation}
Absolute approximation error therefore vanishes with the horizon, even though its share of the squared coefficient size remains positive when $h<\operatorname{rank}(B_{\infty,p})$.

\section{Choosing Patch Length}
\label{sec:selection}

Choosing a patch length from finite data also requires accounting for estimation error. We connect the population results to the finite-window target, establish an estimation bound, and derive its implications for patch choice. Throughout, $L$, $H$, and $m\ge1$ are fixed.

Truncating an infinite-past forecast need not yield the optimal coefficients for a finite observation window. Under the diagonalizability assumption on $F$ in Section~\ref{sec:model}, the finite-window argument underlying Theorem~\ref{thm:bridge} in Appendix~\ref{sec:estimation} gives, for fixed $H$ and any fixed $\bar\rho_F\in(\rho_F,1)$,
\begin{equation}
 \|B_p^\circ-B_{H,p}\|_\Gamma=O(\bar\rho_F^L),
\end{equation}
uniformly over admissible patch lengths. Since $\rho_F<1$, the difference becomes geometrically negligible as $L$ increases. In particular, when $m\ge r$, Theorem~\ref{thm:factor} gives $A_m(p)=O(\bar\rho_F^{2L})$, without requiring exact low rank of $B_p^\circ$.

\subsection{Estimation from Overlapping Windows}

To account for estimation, let $\widehat B_{p,m,H}$ be a global least-squares estimator over $\mathcal{M}_m(p)$, fitted to the same $n$ consecutive windows for each $p$ in the finite nonempty set $\mathcal{P}_L$ of candidate divisors of $L$. Use the same windows for every candidate. The empirical counterpart of the population approximation problem is
\begin{equation}
 \widehat B_{p,m,H}\in
 \operatorname*{arg\,min}_{B\in\mathcal{M}_m(p)}
 \frac1n\sum_{t=1}^n
 \{X_{t+H}-\langle B,Z_{t,p}\rangle_{\mathrm{F}}\}^2.
 \label{eq:patch-empirical-fit}
\end{equation}
For an independent stationary test window, the conditional prediction risk of this fitted coefficient is $R_0+\|\widehat B_{p,m,H}-B_p^\circ\|_\Gamma^2$, by the projection identity in Section~\ref{sec:setup}. It therefore suffices to control this prediction-norm error.

The estimation bound depends on the dimension of the coefficient class $\mathcal{M}_m(p)$, denoted by
\begin{equation}
 d_m(p)=
 \begin{cases}
 m(L/p+p-m), & m<\min\{L/p,p\},\\
 L, & m\ge\min\{L/p,p\}.
 \end{cases}
 \label{eq:dimension}
\end{equation}

\begin{theorem}[Finite-sample prediction bound]
\label{thm:patch-estimation}
Maintain the stationary state-space assumptions of Section~\ref{sec:model}, and additionally assume Gaussian process and measurement noises. There exist positive constants $C_0,C_1,C_2$, depending only on the state-space model, $H$, and $L$, such that, for every $0<\delta<1$, if
\begin{equation}
 n\ge C_0L\max_{p\in\mathcal{P}_L}
 \left\{
 d_m(p)+m\log\frac{12|\mathcal{P}_L|}{\delta}
 \right\},
 \label{eq:patch-estimation-sample-size}
\end{equation}
then, with probability at least $1-\delta$, simultaneously for every $p\in\mathcal{P}_L$, any global minimizer in \eqref{eq:patch-empirical-fit} satisfies
\begin{align}
&\|\widehat B_{p,m,H}-B_p^\circ\|_\Gamma^2 \nonumber\\
&\quad\le C_1 A_m(p)+\frac{C_2}{n}
 \left\{
 d_m(p)+m\log\frac{12|\mathcal{P}_L|}{\delta}
 \right\}. \label{eq:patch-estimation-bound}
\end{align}
The constants are common to all patch lengths and independent of $n$, $m$, $\mathcal{P}_L$, and $\delta$. The bound applies whether or not the rank constraint is active.
\end{theorem}

In \eqref{eq:patch-estimation-bound}, $n$ counts the observed training windows. Their overlap and serial dependence are accounted for in the constants and the sample-size requirement.  The Gaussian assumption in Theorem~\ref{thm:patch-estimation} can be relaxed under suitable independence and tail conditions on the driving noise. Sub-Gaussian and sub-exponential extensions require adapting the concentration bounds in Appendix~\ref{app:dependent-rate}, while retaining the deterministic oracle argument. The constants, rates, and sample-size requirements may change. The result is a prediction-error upper bound, not an exact bias--variance decomposition, and does not require the constrained estimator to be unbiased.

\subsection{Patch Selection and the Approximation--Estimation Tradeoff}

For fixed $L$, $H$, $m$, and confidence level, the logarithmic term in \eqref{eq:patch-estimation-bound} is common to all candidates. Comparing the stated prediction bounds therefore reduces to minimizing
\begin{equation}
 C_1 A_m(p)+\frac{C_2}{n}d_m(p),
 \qquad p\in\mathcal{P}_L,
 \label{eq:proxy}
\end{equation}
where $C_1$ and $C_2$ are the constants in Theorem~\ref{thm:patch-estimation}. Since the approximation errors and these constants may be unknown, \eqref{eq:proxy} is an oracle comparison of prediction bounds, rather than a feasible selection rule or an exact expression for prediction risk.

When the rank constraint is active, $m<\min\{L/p,p\}$, minimizing $d_m(p)$ for fixed $L$ and $m$ is equivalent to minimizing $p+L/p$, which favors a balanced choice of patch length and patch count, with $p=L/p=\sqrt L$ when admissible. Balanced patches therefore minimize the estimation penalty in \eqref{eq:proxy}, but need not minimize the approximation error $A_m(p)$.

\begin{theorem}[Balanced patch lengths]
\label{thm:patch-selection}
Maintain the conditions of Theorem~\ref{thm:patch-estimation}, including \eqref{eq:patch-estimation-sample-size}, and suppose $m<\min\{L/q,q\}$ for every $q\in\mathcal{P}_L$. Let $p\in\mathcal{P}_L$ minimize $d_m(p)$. If
\begin{equation}
 C_1\{A_m(p)-A_m(q)\}
 <\frac{C_2}{n}\{d_m(q)-d_m(p)\}
 \label{eq:patch-selection-margin}
\end{equation}
for every $q\in\mathcal{P}_L$ with $d_m(q)>d_m(p)$, then every minimizer of \eqref{eq:proxy} minimizes $p+L/p$ over $\mathcal{P}_L$. In particular, if $\sqrt L\in\mathcal{P}_L$, it uniquely minimizes \eqref{eq:proxy}.
\end{theorem}

Condition~\eqref{eq:patch-selection-margin} allows a less balanced patch to approximate the target better, provided that this improvement is smaller than its additional estimation penalty. If $A_m(p)$ is constant across candidates, as when $\operatorname{rank}(B_p^\circ)\le m$ for every candidate, the criterion and $d_m(p)$ have exactly the same minimizers. If several divisors attain the minimum dimension, differences in approximation error determine their ordering. The margin shrinks as $n$ grows: for two candidates with $d_m(p)<d_m(q)$ but $A_m(p)>A_m(q)$, the criterion at $p$ is no larger than that at $q$ precisely when
\begin{equation}
 n\le\frac{C_2}{C_1}
 \frac{d_m(q)-d_m(p)}{A_m(p)-A_m(q)}.
 \label{eq:patch-sample-transition}
\end{equation}
Equality gives a tie, and the comparison applies only to sample sizes satisfying \eqref{eq:patch-estimation-sample-size}. The smaller class is favored below this threshold and the more accurate representation above it. The square-root rule is therefore not universal: its advantage comes from estimation.

\paragraph{Extreme and intermediate patch lengths.}
The two extreme patch lengths, point tokens ($p=1$) and a single token containing the whole window ($p=L$), lie at one end of the tradeoff in \eqref{eq:proxy}. Their coefficient matrix is a single column or row, so its rank is at most one and one head represents every linear forecast: $A_m(p)=0$ for any target. The price is that the rank constraint is inactive. Nothing is shared across patches, the fitted class is the unrestricted set of linear forecasts on the window, and $d_m(p)=L$, the largest possible dimension. Rank one is automatic here and is not a form of compression (Lemma~\ref{lem:patch-endpoints}). An intermediate patch length with $m<\min\{L/p,p\}$ makes the opposite trade. All patches must reuse the same $m$ within-patch weighting patterns, which lowers the dimension to $d_m(p)=m(L/p+p-m)<L$. This sharing can introduce approximation error, since $A_m(p)>0$ unless $\operatorname{rank}(B_p^\circ)\le m$. Taking $q$ to be an extreme patch length in \eqref{eq:patch-sample-transition} shows that the criterion is no larger at the intermediate patch exactly when $C_1A_m(p)\le C_2\{L-d_m(p)\}/n$, that is, when the approximation error it adds is at most the estimation penalty it saves. This favors intermediate patches when data are limited or the reshaped target is close to rank $m$. Because the saving shrinks with $n$, an extreme patch length is eventually preferred whenever $A_m(p)>0$. Computational cost, which the bound omits, also falls as the token count $L/p$ decreases (Appendix~\ref{app:cost-details}).

\section{NONLINEAR DECODING AND SCOPE}
\label{sec:nonlinear-scope}

Although our guarantees concern a structured attention model, they isolate mechanisms relevant to more general patch-based forecasters. Patch length changes how temporal information is grouped and how representations are shared, affecting both predictive approximation and learning complexity. The distinction between forecasting and filtering also explains why a longer horizon need not simplify extracting predictive information from the past. These insights motivate tuning patch length jointly with model capacity and using balanced patches as a reference when representation errors are comparable. More expressive architectures introduce additional effects, but the underlying representation--estimation tradeoff remains a useful organizing principle.

\begin{proposition}[Nonlinear decoding after shared patch encoding]
\label{prop:nonlinear-patch-encoding}
Under the notation of Section~\ref{sec:setup}, suppose
$(x_{t,L},X_{t+H})$ is jointly Gaussian. For a shared
encoder $E\in\mathbb{R}^{p\times m}$ and any measurable
decoder $g$ with square-integrable output,
\begin{equation}
\inf_{E,g}
\mathbb{E}\!\left[
  \{X_{t+H}-g(Z_{t,p}E)\}^{2}
\right]
=R_0+A_m(p).
\end{equation}
Here $m$ is the encoder width, and the decoder receives
only the encoded patches $Z_{t,p}E$.
\end{proposition}

Thus the population approximation limit also applies to nonlinear attention and multilayer decoders operating only on the encoded patches. This result concerns encoder width. It does not extend the head-count characterization or the estimation bound to arbitrary nonlinear models. The proof is in Appendix~\ref{app:nonlinear-decoding}.

\section{SIMULATIONS}
\label{sec:experiments}

\begin{figure*}[!t]
 \centering
 \includegraphics[width=0.77\textwidth]{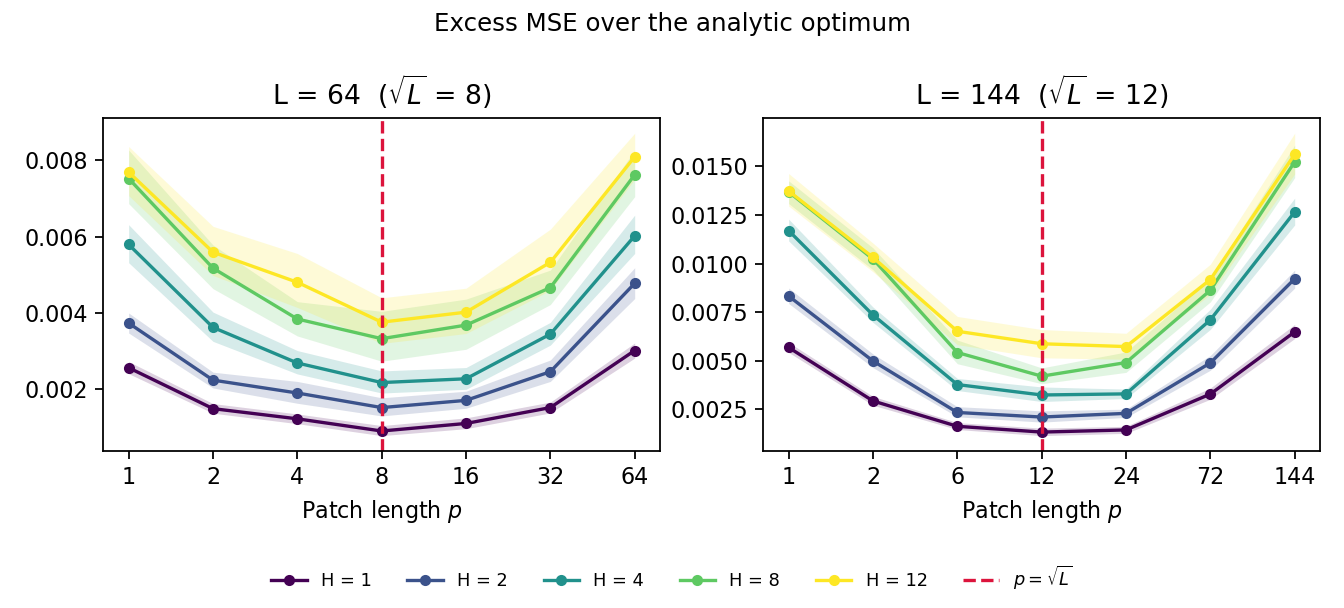}
\caption{Excess MSE across forecast horizons for fixed one-head attention, with $n=200$ independent training series. Curves show means over $50$ data repetitions. Shaded bands are pointwise $95\%$ bootstrap intervals. Dashed lines mark $p=\sqrt L$. Patch candidates are displayed at equally spaced categorical positions, and panel scales differ.}
 \label{fig:simulation-horizons}
\end{figure*}

\begin{figure*}[!t]
 \centering
 \includegraphics[width=0.77\textwidth]{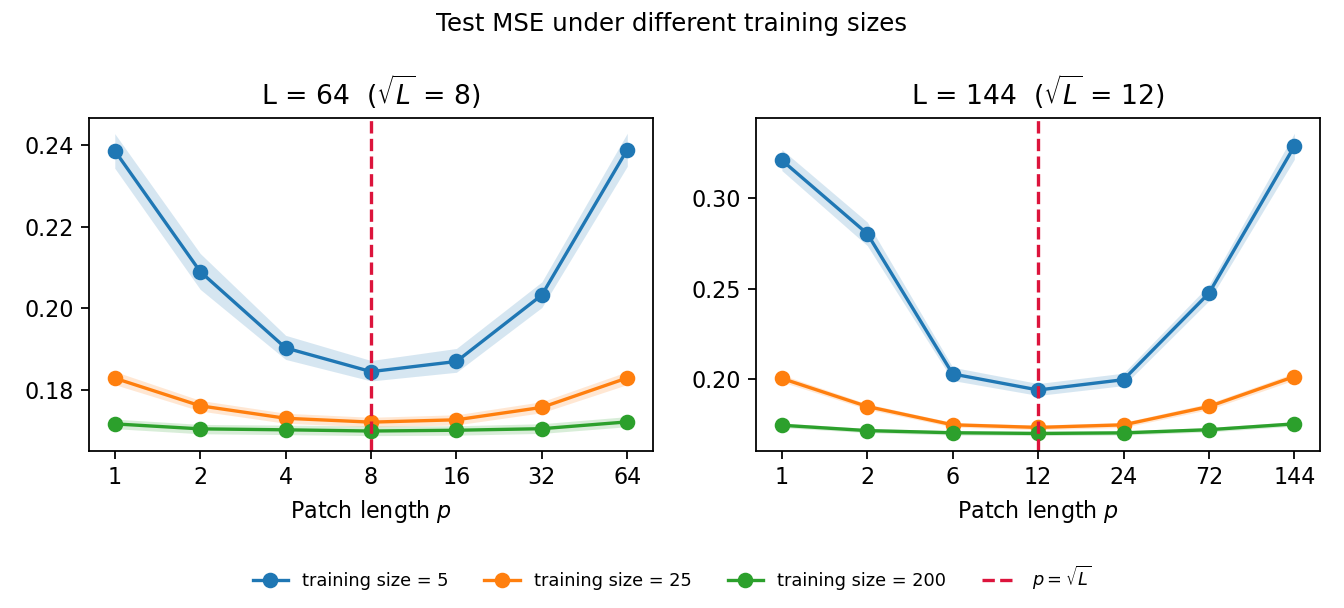}
\caption{Total test MSE at $H=1$ for fixed one-head attention as the number of independent training series grows. Means and pointwise $95\%$ bootstrap intervals use $50$ repetitions. Other plotting conventions match Figure~\ref{fig:simulation-horizons}.}
 \label{fig:simulation-training}
\end{figure*}

\subsection{Design}

We generate stationary paths from a three-state model of the form \eqref{eq:ssm}, whose parameters are given in Appendix~\ref{app:sim-design}. The forecasting target is $X_{t+H}$, using the latest $L$ observations. We use window lengths $L\in\{64,144\}$, horizons $H\in\{1,2,4,8,12\}$, and $n\in\{5,25,200\}$ independent training series.

The fitted model is one-head attention with fixed attention weights: a constant query, a learnable key for each patch position, and the same linear value map for every patch. It can represent every rank-one coefficient matrix, so the experiment isolates how patch length affects learning. Appendix~\ref{app:experiments} gives the design details, the population construction, diagnostic definitions, and reporting protocol.

\subsection{Forecast Horizon and the Interior Patch Range}

Figure~\ref{fig:simulation-horizons} reports excess MSE: the held-out mean of $(\widehat f(x)-f^\circ_{H,L}(x))^2$, where $f^\circ_{H,L}$ is the analytic finite-window conditional mean. At the population level, total forecast risk is the sum of this excess risk and the Bayes residual variance. Measuring the distance to the conditional mean removes the unpredictable noise in the target, which would otherwise hide differences between fitted models.

Across the displayed horizons, the curves have a broad interior minimum near $\sqrt L$, with larger errors at both extremes. The preferred range moves toward larger patches when the window grows, consistent with the dimension balance in \eqref{eq:dimension}. However, neighboring patches remain competitive: at the longest horizon, a larger neighboring patch can match or slightly improve on the balanced reference, without a statistically resolved difference. The evidence therefore favors a range of patch lengths rather than a unique optimum.

Fitted excess error also increases across these horizons. A more distant target does not automatically make the one-head forecasting problem easier, even as its predictable signal becomes smaller. Appendix~\ref{app:add-floor} reports the population diagnostic for the persistent relative filtering error predicted by Theorem~\ref{thm:floor}. The fitted curves additionally reflect finite data and the training procedure.

\begin{table}[!t]
 \centering
\caption{One-step test MSE in units of $10^{-3}$ for the balanced patch and both extremes. Entries are means $\pm$ standard errors over $50$ data repetitions, and $n$ counts independent training series.}
 \label{tab:simulation-main}
 \begin{tabular}{@{}rrccc@{}}
 \toprule
 $L$ & $n$ & $p=1$ & $p=\sqrt L$ & $p=L$\\
 \midrule
 64 & 5 & $238.6 \pm 2.2$ & \textbf{$184.4 \pm 1.3$} & $238.9 \pm 2.1$\\
 64 & 25 & $182.8 \pm 0.8$ & \textbf{$172.0 \pm 0.6$} & $182.8 \pm 0.8$\\
 64 & 200 & $171.6 \pm 0.6$ & \textbf{$169.8 \pm 0.6$} & $172.0 \pm 0.6$\\
 \midrule
 144 & 5 & $320.7 \pm 3.0$ & \textbf{$194.1 \pm 1.7$} & $328.4 \pm 3.7$\\
 144 & 25 & $200.5 \pm 1.0$ & \textbf{$173.5 \pm 0.7$} & $201.5 \pm 1.0$\\
 144 & 200 & $174.7 \pm 0.6$ & \textbf{$170.2 \pm 0.6$} & $175.4 \pm 0.6$\\
 \bottomrule
 \end{tabular}
\end{table}

\subsection{Training Size and Interpretation}

Figure~\ref{fig:simulation-training} and Table~\ref{tab:simulation-main} show that interior patches have their largest advantage when training data are scarce. As more independent series become available, prediction error falls throughout the grid and the absolute gap between interior and extreme patch lengths narrows. This pattern appears at both window lengths. The preferred interior patch also varies across the displayed horizons, so the balanced reference need not minimize every fitted curve.

The results are consistent with the estimation term in \eqref{eq:proxy} and with the extreme-versus-intermediate comparison in Section~\ref{sec:selection}. Here the approximation error of interior patches is negligible (Appendix~\ref{app:add-repr}), so the disadvantage of the extremes reflects estimation.

The experiment nevertheless does not identify class dimension as the sole explanation: network parameter count is minimized at the same patch length in this architecture, and regularization and optimization can affect the fitted curves. Taken together, the horizon and sample-size comparisons support testing an interior range around $\sqrt L$, with validation used to choose among nearby candidates. Two window lengths alone cannot establish a square-root scaling law, and Appendix~\ref{app:add-scaling} examines nine.

\section{CONCLUSION}
\label{sec:conclusion}

This paper makes three contributions. First, we show that multi-head patch attention with shared linear values forecasts through a coefficient matrix of rank at most the head count, even with input-dependent weights; for Gaussian data, no decoder on a width-$m$ shared linear patch encoding can do better than the best rank-$m$ matrix. Second, for linear state-space models, an exact factorization shows that a dominant long-horizon component need not be rank-one representable. Third, with dependent-data estimation bounds, we derive a patch-selection criterion that balances representation and estimation errors and identifies when balanced patches are justified. Future work includes learned content-dependent attention.

\bibliography{references}

\clearpage
\appendix
\thispagestyle{empty}

\onecolumn
\aistatstitle{Patch Attention Wisely: Understanding Patch Length in Attention-Based Forecasting \\
Supplementary Materials}

\section{GUIDE AND NOTATION}
\label{app:roadmap}

The appendix follows the argument of the main paper: attention representation, state-space approximation, finite-window prediction, estimation from dependent data, and patch selection. Each result is followed by its proof. The remaining sections treat context adaptation, practical implications, and the simulations.

\begin{itemize}
\item Appendix~\ref{app:representation} proves the coefficient-matrix identity and gives explicit fixed positional attention constructions for Section~\ref{sec:setup}.
\item Appendix~\ref{app:spectral-proofs} proves the filtering-mode factorization, the long-horizon floor, and the shared-horizon extension from Section~\ref{sec:model}.
\item Appendix~\ref{sec:estimation} relates the truncated infinite-past coefficients to the optimal finite-window forecast.
\item Appendix~\ref{app:dependent-rate} derives the dependent-data rate and proves Theorem~\ref{thm:patch-estimation}.
\item Appendix~\ref{app:patch-bounds} states a conditional selection criterion, proves the patch-selection results, and collects the dimension and computational-cost calculations.
\item Appendix~\ref{app:context-mixture-details} studies adaptation across generating systems.
\item Appendix~\ref{sec:discussion-implications} discusses when context dependence can help and how patch lengths should be compared in practice.
\item Appendix~\ref{app:experiments} gives the simulation diagnostics and evaluation protocol.
\item Appendix~\ref{app:additional-experiments} reports additional experiments: a second observation design, the population filtering floor, head counts for rank-one and rank-three targets, and nine window lengths.
\end{itemize}

\subsection{Notation and types of results}
\label{app:notation}

We keep the notation from the main paper. The context contains $L$ observations, each patch contains $p$ observations, and $N=L/p$ is the number of patches. The forecast horizon is $H$, and $h$ denotes a rank or head budget. The main paper uses $m$ for the fixed head budget. The matrix $B_{H,p}$ is obtained by truncating the infinite-past forecast to the latest $L$ observations and reshaping its coefficients into $N$ rows and $p$ columns. The matrix $B^\circ_{H,L,p}$ instead denotes the best coefficient matrix when the forecaster is allowed to use only those $L$ observations. These matrices are close for a long stable context, but they need not be identical.

For a coefficient error $D$, the prediction norm $\|D\|_{\Gamma_{L,p}}^2$ is the mean squared change in the forecast caused by $D$. It is the relevant error measure because it accounts for correlation among lagged observations. When ordinary Frobenius or operator norms are used instead, the text says so explicitly. All additional symbols are introduced in the section where they are needed.

The supplement distinguishes three kinds of statements. Representation results describe which fixed forecast coefficients an attention model can express. Population results describe the best predictor when the data-generating system is known. Estimation results describe what can be learned from a finite, dependent trajectory. Keeping these levels separate is important: a low-rank coefficient matrix does not by itself guarantee that a trained attention model will find it.

\section{ATTENTION REPRESENTATION}
\label{app:representation}

This section establishes the representation claim used in Section~\ref{sec:setup}. The coefficient-matrix identity gives the rank upper bound. The fixed positional construction gives the converse. The subsequent extensions treat shared horizons, signed softmax pooling, approximation, and input-dependent attention weights.

\subsection{Coefficient-matrix identity}
\label{app:context-attention}

Let $Z\in\mathbb{R}^{N\times p}$ contain the raw patches as rows. For head $a$, write its score for patch $j$ as
\begin{equation}
 \alpha_{H,a,j}(Z)=N^{-1}k_{j,a}(Z)^\top q_{H,a}(Z).
\end{equation}
Unlike softmax weights, these scores may be negative and need not sum to one. The factor $1/N$ is only a scaling convention. When models with different token counts are compared, the convention should be kept fixed. The shared linear value path reduces to a vector $b_a\in\mathbb{R}^p$ that tells the head how to weight observations within every patch. Thus head $a$ contributes $\alpha_{H,a}(Z)b_a^\top$, an outer product. Define the resulting coefficient matrix and forecast by
\begin{equation}
 B_H(Z)=\sum_a\alpha_{H,a}(Z)b_a^\top,
 \qquad
 \widehat f_H(Z)=\langle B_H(Z),Z\rangle_{\mathrm{F}}.
\end{equation}

\begin{proof}[Proof of Proposition~\ref{prop:routing}]
With $b_a=W_{V,a}^\top w_a$, expand the forecast:
\begin{equation}
 \widehat f_H(Z)=\sum_{a,j}\alpha_{H,a,j}(Z)b_a^\top x_j
 =\left\langle\sum_a\alpha_{H,a}(Z)b_a^\top,Z\right\rangle_{\mathrm{F}}.
\end{equation}
Each summand is an $N\times p$ outer product, so its sum has rank at most $\min\{m,N,p\}$. Vector-valued values do not change this conclusion for this one scalar linear output, because $W_{V,a}$ and $w_a$ are fixed and shared across positions. Nonlinear values, query-dependent output maps, and other paths need separate treatment. No converse head lower bound for adaptive functions follows from this pointwise identity.
\end{proof}

\subsection{Exact fixed positional pooling}
\label{app:attention-bridge}

\label{app:setup-proofs}

Let a context of length $L=Np$ be divided into nonoverlapping patches $x_j\in\mathbb{R}^p$, and stack $x_j^\top$ as the rows of $X\in\mathbb{R}^{N\times p}$.  For $B\in\mathbb{R}^{N\times p}$, define
\begin{equation}
 f_B(X)=\langle B,X\rangle_{\mathrm{F}}=\sum_{j=1}^N B_{j,:}x_j.
 \label{eq:attn-bridge-target}
\end{equation}
Use $N$ data tokens and one fixed forecast-query token,
\begin{equation}
 z_j=\begin{bmatrix}x_j\\e_j\\0\end{bmatrix},\qquad
 z_\star=\begin{bmatrix}0_p\\0_N\\1\end{bmatrix},\qquad
 D=p+N+1,
 \label{eq:attn-bridge-tokens}
\end{equation}
where $e_j$ is the $j$th absolute positional basis vector.  A permutation-symmetric hard mask admits all $N$ data positions and excludes the forecast token itself. This is equivalent to assigning the forecast token a $-\infty$ softmax logit. A finite mask penalty changes the softmax normalizer and is outside the exact result.

For one scalar unnormalized pooling head, put
\begin{equation}
 o_a(X)=\frac1N\sum_{j=1}^N
 (W_Q^{(a)}z_\star)(W_K^{(a)}z_j)(W_V^{(a)}z_j).
 \label{eq:attn-bridge-head}
\end{equation}
For $m$ heads, stack $o=(o_1,\ldots,o_m)^\top$.  A standard multi-head output projection and scalar output layer have the form
\begin{equation}
 y_\star=W_O o,\qquad \widehat y=r^\top y_\star,\qquad
 W_O\in\mathbb{R}^{D\times m},\quad r\in\mathbb{R}^D.
 \label{eq:attn-bridge-readout}
\end{equation}
Any desired head-mixing vector $c\in\mathbb{R}^m$ is implemented by $W_O=e_1c^\top$ and $r=e_1$, giving $\widehat y=c^\top o$.

\begin{theorem}[Exact fixed positional linear pooling]
\label{thm:attention-linear-exact}
Suppose $\operatorname{rank}(B)\le h$ and choose
\begin{equation}
 B=\sum_{a=1}^h u_av_a^\top,\qquad u_a\in\mathbb{R}^N,\quad v_a\in\mathbb{R}^p.
 \label{eq:attn-bridge-factor}
\end{equation}
There is a one-layer fixed positional pooling construction with $N+1$ tokens, width $D=p+N+1$, $h$ heads, scalar query/key/value dimensions, a hard mask, standard $W_O$, and a scalar output layer that equals $f_B(X)$ for every $X$. Head $a$ uses
\begin{equation}
 W_Q^{(a)}z_\star=1,\qquad
 W_K^{(a)}z_j=N u_{a,j},\qquad
 W_V^{(a)}z_j=v_a^\top x_j,
 \label{eq:attn-bridge-linear-projections}
\end{equation}
and \eqref{eq:attn-bridge-readout} uses $c=\mathbf 1_h$.
\end{theorem}

\begin{proof}
Head $a$ returns $o_a(X)=N^{-1}\sum_j(Nu_{a,j})v_a^\top x_j=\langle u_av_a^\top,X\rangle_{\mathrm{F}}$. The positional key factor $N$ exactly cancels the fixed token-count factor. Summing the heads and using \eqref{eq:attn-bridge-factor} gives $f_B(X)$.
\end{proof}

\begin{remark}[Positional parameters]
The explicit width uses $N$ one-hot positional coordinates.  A $B$-specific learned positional table may instead store $(u_{1,j},\ldots,u_{h,j})\in\mathbb{R}^h$, reducing width to $p+h+1$ while retaining $Nh$ positional parameters.  The smaller width does not eliminate the order information needed to store a general $N\times h$ factor.
\end{remark}

\begin{proposition}[Restricted-architecture head lower bound]
\label{prop:attention-head-lower}
Consider one hard-masked layer with a fixed forecast query, scalar values that are fixed linear functionals of $x_j$, content-independent scalar keys that depend only on position, standard $W_O$ plus a scalar linear output layer, and no residual or feed-forward bypass.  Every such $m$-head unnormalized pooling layer represents a coefficient matrix of rank at most $m$.  Thus $m\ge\operatorname{rank}(B)$ is necessary within this full restricted architecture.
\end{proposition}

\begin{proof}
After absorbing the scalar query and output coefficient, head $a$ contributes some rank-one matrix $u_av_a^\top$.  The sum of $m$ such matrices has rank at most $m$.  This fixed-operator lower bound does not extend to arbitrary content-dependent functions, nonlinear values, depth, or another input-to-output path. Vector-valued linear values alone do not remove the rank bound under the shared scalar output weights of Proposition~\ref{prop:routing}.
\end{proof}

\subsection{Shared pooling across finitely many horizons}

\begin{corollary}[Finite shared multi-horizon pooling]
\label{cor:attention-multihorizon}
Let $B^{(t)}\in\mathbb{R}^{N\times p}$ for $t=1,\ldots,T$ satisfy
\begin{equation}
 B^{(t)}=\sum_{a=1}^h d_{t,a}u_av_a^\top.
 \label{eq:attn-bridge-cp}
\end{equation}
Use $N$ data tokens and $T$ one-hot horizon-query tokens, all of width $D=p+N+T$.  One hard-masked layer with $h$ scalar heads and shared key/value projections outputs $f_{B^{(t)}}(X)$ at query $t$ by setting
\begin{equation}
 W_Q^{(a)}z_\star^{(t)}=d_{t,a},\qquad
 W_K^{(a)}z_j=N u_{a,j},\qquad
 W_V^{(a)}z_j=v_a^\top x_j.
\end{equation}
The construction uses $N+T$ tokens and explicitly stores a $T\times h$ table $(d_{t,a})$.
\end{corollary}

\begin{proof}
The mask admits exactly the $N$ data tokens. At horizon query $t$, head $a$ outputs $N^{-1}\sum_j d_{t,a}(N u_{a,j})v_a^\top x_j$. Summing over heads gives $\langle B^{(t)},X\rangle_{\mathrm{F}}$. The factor $1/N$ counts the admitted data tokens, not the $N+T$ total tokens.
\end{proof}

\begin{remark}[No unseen-horizon extrapolation]
The corollary is transductive over the displayed $T$ horizons.  It provides no coefficient for an unseen horizon.  Such an extension requires a specified feature map $\phi(H)$ and a structural rule, for example $d_{H,a}=\theta_a^\top\phi(H)$.
\end{remark}

\subsection{Signed softmax realization}

\begin{theorem}[Exact signed fixed positional softmax pooling]
\label{thm:attention-softmax-exact}
Under \eqref{eq:attn-bridge-factor}, fix $\epsilon>0$.  Let $u_a^+=(u_a)_+$, $u_a^-=(-u_a)_+$, and
\begin{equation}
 s_a^\pm=\lVert u_a^\pm\rVert_1,\qquad
 c_a^\pm=s_a^\pm+N\epsilon,\qquad
 \ell_{a,j}^\pm=\log(u_{a,j}^\pm+\epsilon).
 \label{eq:attn-bridge-soft-defs}
\end{equation}
A hard-masked fixed positional softmax-pooling layer with $N+1$ tokens, width $p+N+1$, $2h$ scalar heads, standard $W_O$, and a scalar output layer represents $f_B$ exactly.  The $(a,\pm)$ head uses query $1$, positional logit $\ell_{a,j}^\pm$, and value $v_a^\top x_j$.  The head-mixing vector in \eqref{eq:attn-bridge-readout} is $(c_1^+,-c_1^-,\ldots,c_h^+,-c_h^-)$.
\end{theorem}

\begin{proof}
The $(a,\pm)$ attention probability is
\begin{equation}
 \pi_{a,j}^\pm=\frac{u_{a,j}^\pm+\epsilon}{s_a^\pm+N\epsilon}.
\end{equation}
The signed output for pair $a$ is therefore
\begin{equation}
 c_a^+\sum_j\pi_{a,j}^+v_a^\top x_j
 -c_a^-\sum_j\pi_{a,j}^-v_a^\top x_j
 =\sum_j u_{a,j}v_a^\top x_j,
\end{equation}
because the two $\epsilon$ offsets cancel.  Summing over $a$ proves the claim.
\end{proof}

\begin{remark}[Architecture and conditioning restrictions]
The theorem fixes one query, a permutation-symmetric hard mask, absolute positional keys, scalar patch-linear values, tied values within each sign pair, one layer, no bypass, and signed head mixing through $W_O$ and a scalar output layer.  Moreover, its numerical conditioning depends on the chosen factorization and $\epsilon$: rescaling $u_a\mapsto\alpha_a u_a$ and $v_a\mapsto v_a/\alpha_a$ leaves $B$ fixed but changes logits, output weights, and perturbation sensitivity.
\end{remark}

\subsection{Approximation and perturbation bounds}

Let $B_h$ be a best rank-$h$ Frobenius approximation to $B$.

\begin{theorem}[Approximation and covariance-weighted prediction error]
\label{thm:attention-approximation}
Either exact construction applied to $B_h$ gives a fixed positional pooling output $f_{B_h}(X)$ such that
\begin{align}
|f_B(X)-f_{B_h}(X)|
 &\le \lVert B-B_h\rVert_{\mathrm{F}}\lVert X\rVert_{\mathrm{F}}, \nonumber\\
\lVert B-B_h\rVert_{\mathrm{F}}^2
 &=\sum_{k>h}\sigma_k(B)^2. \label{eq:attn-bridge-tail}
\end{align}
If $\Gamma=\mathbb E[\operatorname{vec}(X)\operatorname{vec}(X)^\top]$, then
\begin{equation}
 \mathbb E\{f_B(X)-f_{B_h}(X)\}^2
 \le\lambda_{\max}(\Gamma)\sum_{k>h}\sigma_k(B)^2.
 \label{eq:attn-bridge-risk}
\end{equation}
\end{theorem}

\begin{proof}
The first statement is Cauchy--Schwarz and Eckart--Young--Mirsky.  Vectorizing the Frobenius inner product gives the covariance-weighted statement.
\end{proof}

\begin{lemma}[Softmax perturbation]
\label{lem:attention-softmax-perturbation}
For $r,s\in\mathbb{R}^N$, $\lVert\operatorname{softmax}(r)-\operatorname{softmax}(s)\rVert_1 \le2\lVert r-s\rVert_\infty$.  Consequently, if every signed logit in Theorem~\ref{thm:attention-softmax-exact} is approximated within $\delta$ in $\ell_\infty$, the induced coefficient matrix $\widetilde B_h$ obeys
\begin{equation}
 \lVert\widetilde B_h-B_h\rVert_{\mathrm{F}}
 \le2\delta\sum_{a=1}^h(c_a^++c_a^-)\lVert v_a\rVert_2.
 \label{eq:attn-bridge-logit-error}
\end{equation}
\end{lemma}

\begin{proof}
Along the segment between $r$ and $s$, the softmax Jacobian bounds the $\ell_\infty$-to-$\ell_1$ derivative norm by $2$.  Each head coefficient perturbation is an outer product of the changed probability vector and $v_a$. The triangle inequality over the $2h$ heads proves \eqref{eq:attn-bridge-logit-error}.
\end{proof}

\subsection{The role of positional information}

\begin{proposition}[Order information under permutation-symmetric pooling]
\label{prop:attention-order-information}
Fix a forecast query and a mask that treats all admitted data positions symmetrically.  If every token is formed by the same map $z_j=g(x_j)$ with no absolute, relative, causal-distance, convolutional, or other order-dependent information, the pooling output is permutation invariant.  A linear $f_B$ is permutation invariant for every $X$ only if all rows of $B$ are equal. Therefore a general coefficient matrix needs some symmetry-breaking order information under these assumptions.
\end{proposition}

\begin{proof}
The pooling operation depends only on the multiset of key--value pairs. Invariance of $f_B$ under arbitrary row permutations forces every pair of rows of $B$ to agree. The proposition does not privilege absolute one-hot positions and does not apply to asymmetric masks or content maps that already encode order.
\end{proof}

\paragraph{Relation to existing attention theory.} Goel and Bartlett study causal softmax Transformers that approximate Kalman filtering \citep{goel2024transformer}, Sander et al. study autoregressive learning in context \citep{sander2024autoregressive}, and Lu and Yang connect linear attention to dynamic VAR computations \citep{lu2025var}. Here the object is a fixed finite-window coefficient matrix obtained by a nonoverlapping reshape. Page-matrix and spectral-LDS methods provide related representations of time-series predictors \citep{damen1982page,rashidinejad2020slip}. The constructions above specify how the positional tokens, mask, query/key/value maps, and scalar output layer realize that matrix. They do not provide recursive filtering or optimization guarantees.

\subsection{Input-dependent attention weights and the full Jacobian}

All learned parameters are held fixed when taking derivatives or population expectations. The following results distinguish the pointwise coefficient matrix from the derivative of the complete forecast.

\begin{proposition}[Attention weights and the full input Jacobian]
\label{prop:adaptive-jacobian}
At differentiability points of $B_H$, the Frobenius gradient is
\begin{equation}
 J_H(Z)=\nabla_Z\widehat f_H(Z)
 =B_H(Z)+\sum_{a,j}(b_a^\top x_j)\nabla_Z\alpha_{H,a,j}(Z).
 \label{eq:adaptive-jacobian}
\end{equation}
Only the first term is the derivative obtained by freezing the attention weights at the current input. The second need not vanish or preserve the rank bound.
\end{proposition}
\begin{proof}
Apply the product rule to each term $\alpha_{H,a,j}(Z)b_a^\top x_j$ and sum.
\end{proof}

\paragraph{Why the Jacobian rank can be larger.} To see the distinction algebraically, use one-based indices, let one head have $b=(1,0)^\top$, take scalar keys $k_j=x_{j,2}$, and fix the scalar query. Then
\begin{equation}
 \alpha_j=N^{-1}x_{j,2},\qquad
 f(Z)=N^{-1}\sum_{j=1}^N x_{j,1}x_{j,2}.
\end{equation}
The coefficient matrix has only its first column nonzero and therefore has rank at most one. By contrast, the first two columns of the input Jacobian are $N^{-1}Z[:,2]$ and $N^{-1}Z[:,1]$. They are generically linearly independent, so the Jacobian can have rank two. Thus content-dependent keys can increase sensitivity rank even with a fixed query and a rank-one coefficient matrix.

\paragraph{Three different matrices.} The unmasked token mixing matrix $N^{-1}\mathbf Q\mathbf K^\top$ has shape $N\times N$ and rank at most $\min(N,d_k)$. It is different from the $N\times p$ coefficient matrix $B_H(Z)$ and the $N\times p$ input gradient $J_H(Z)$. A triangular causal mask can invalidate the token-matrix rank bound. None of these statements bounds the number of trainable parameters in an arbitrary context encoder.

\subsection{Comparison of adaptive attention weights with a fixed benchmark}

\begin{proposition}[Comparison with a fixed spectral benchmark]
\label{prop:adaptive-static-bound}
Let $B_0$ be deterministic, let $B_{0,h}$ be its rank-$h$ truncated SVD, and let $\Gamma=\mathbb E[\operatorname{vec}(Z)\operatorname{vec}(Z)^\top]$. For the payload-linear adaptive model, when the right side is finite,
\begin{align}
\mathbb E|\widehat f_H(Z)-\langle B_0,Z\rangle_{\mathrm{F}}|^2
 &\le 2\lambda_{\max}(\Gamma)\sum_{k>h}\sigma_k(B_0)^2 \nonumber\\
&\quad+2\mathbb E\{\|B_H(Z)-B_{0,h}\|_{\mathrm{F}}^2\|Z\|_{\mathrm{F}}^2\}. \label{eq:adaptive-static-bound}
\end{align}
\end{proposition}
\begin{proof}
Decompose the prediction error into $\langle B_H(Z)-B_{0,h},Z\rangle_{\mathrm{F}}+\langle B_{0,h}-B_0,Z\rangle_{\mathrm{F}}$. Use $(u+v)^2\le2u^2+2v^2$, pointwise Cauchy--Schwarz for the first term, and the deterministic covariance bound for the second.
\end{proof}
The second term contains representation and fitting error. The singular tail alone cannot control it. Since $B_H(Z)$ depends on $Z$, replacing this term by an unconditional covariance eigenvalue times its expected squared matrix norm is unjustified without an additional condition. Taking $B_0=B^\circ_{H,L,p}$ compares to the finite-window oracle. Taking $B_0=B_{H,p}$ compares to the truncated infinite-past kernel and still requires the finite-window bridge. This proposition supplies no sample-size rate for learned attention.

\subsection{Nonlinear decoding after shared patch encoding}
\label{app:nonlinear-decoding}

\begin{proof}[Proof of Proposition \ref{prop:nonlinear-patch-encoding}]
For fixed $E$, Gaussian conditioning implies that the
optimal decoder is linear:
$\mathbb{E}[X_{t+H}\mid Z_{t,p}E]
=\langle D,Z_{t,p}E\rangle_{\mathrm{F}}$
for some $D\in\mathbb{R}^{N\times m}$. Since
\[
\langle D,Z_{t,p}E\rangle_{\mathrm{F}}
=\langle DE^\top,Z_{t,p}\rangle_{\mathrm{F}},
\]
and the products $DE^\top$ range over all matrices of
rank at most $m$, the projection identity \eqref{eq:risk} yields
\[
\inf_{E,g}\mathbb{E}
\{X_{t+H}-g(Z_{t,p}E)\}^{2}
=
R_0+\inf_{\operatorname{rank}(B)\le m}
\|B-B_p^\circ\|_\Gamma^2.
\]
The last infimum is $A_m(p)$.
\end{proof}

\section{STATE-SPACE FILTERING AND APPROXIMATION}
\label{app:spectral-proofs}

The state-space calculation separates two jobs. The transition matrix $A$ propagates the current state toward the forecast horizon, while the filtering matrix $F$ describes how past observations reveal that state. The proofs below show that patching does not create new filtering modes: it only splits each lag pattern into an across-patch part and a within-patch part.

\subsection{Filtering-mode factorization}

\begin{proof}[Proof of Theorem~\ref{thm:factor}]
The proof has one idea: each exponential lag pattern splits into an across-patch factor and a within-patch factor because $z^{jp+s}=(z^p)^jz^s$. The forecast horizon changes the coefficient of that pattern, not its shape.

By spectral decomposition,
\begin{equation}
A^{H-1}=\sum_{a=1}^r\lambda_a^{H-1}P_a,
\qquad
F^\ell=\sum_{k=1}^r\varphi_k^\ell Q_k.
\end{equation}
Substituting into \eqref{eq:kernel} gives
\begin{equation}
\beta_H(\ell)=\sum_{a=1}^r\sum_{k=1}^r
(CP_aQ_kK)\lambda_a^{H-1}\varphi_k^\ell
=\sum_{k=1}^r d_k(H)\varphi_k^\ell.
\end{equation}
With $\ell=jp+s$,
\begin{equation}
\beta_H(jp+s)=\sum_{k=1}^r d_k(H)(\varphi_k^p)^j\varphi_k^s,
\end{equation}
which is exactly \eqref{eq:factor}. For nonreal modes this identity is over \(\mathbb C\), with conjugate terms summing to the real matrix \(B_{H,p}\). Each summand is rank one over \(\mathbb C\), which proves the stated rank bound. Because \(B_{H,p}\) is real, its real and complex matrix ranks agree.
\end{proof}

\subsection{Long-horizon filtering floor}

\begin{proof}[Proof of Theorem~\ref{thm:floor}]
Since $A$ is real, nonreal eigenvalues occur in conjugate pairs with equal modulus. The uniquely dominant $\lambda_1$ is therefore real. Consequently $B_{\infty,p}$ and the rescaled matrices below are real, even when individual filtering factors are complex.

Theorem~\ref{thm:factor} gives
\begin{equation}
\lambda_1^{-(H-1)}B_{H,p}-B_{\infty,p}
=\sum_{a=2}^r\sum_{k=1}^r
g_{ak}\left(\frac{\lambda_a}{\lambda_1}\right)^{H-1}
u_{N,p}(\varphi_k)v_p(\varphi_k)^\top.
\end{equation}
The triangle inequality and $\|uv^\top\|_{\mathrm{op}}=\|u\|_2\|v\|_2$ prove \eqref{eq:oprate}. Weyl's inequality implies
\begin{equation}
\left|
\frac{\sigma_j(B_{H,p})}{|\lambda_1|^{H-1}}
-\sigma_j(B_{\infty,p})
\right|\le C_pq_A^{H-1}.
\end{equation}
Let $e_H=C_pq_A^{H-1}$, $s_j=\sigma_j(B_{\infty,p})$, and $\widetilde s_j=\sigma_j(\lambda_1^{-(H-1)}B_{H,p})$. If $e_H\le s_1/2$, then $\widetilde s_1\ge s_1/2$ and
\begin{equation}
\left|\frac{\widetilde s_{h+1}}{\widetilde s_1}
-\frac{s_{h+1}}{s_1}\right|
\le \frac{e_H}{\widetilde s_1}
+\frac{s_{h+1}e_H}{s_1\widetilde s_1}
\le\frac{4e_H}{s_1}.
\end{equation}
By Eckart--Young, $R^{\mathrm{op}}_{H,p,h}$ is the relative best rank-$h$ operator-norm error, and \eqref{eq:relrate} follows. If $h<\operatorname{rank}(B_{\infty,p})$, then $s_{h+1}>0$.
\end{proof}

\subsection{Active modes and patch aliasing}

\begin{proof}[Proof of Corollary~\ref{cor:active-modes}]
Let $\mathcal S=\{k:g_{1k}\ne0\}$ be the filtering modes coupled to the dominant forecast component. Restrict the two modal factor matrices to the active set \(\mathcal S\), and write
\begin{equation}
B_{\infty,p}=U_{\mathcal S}
\operatorname{diag}\{g_{1k}:k\in\mathcal S\}
V_{\mathcal S}^\top.
\end{equation}
The columns of \(V_{\mathcal S}\) form a Vandermonde matrix with nodes \(\varphi_k\), and the columns of \(U_{\mathcal S}\) form a Vandermonde matrix with nodes \(\varphi_k^p\). The distinct-node assumptions and \(|\mathcal S|\le\min\{N,p\}\) make both matrices full column rank. The diagonal middle factor is nonsingular, so Sylvester's rank inequality, applied twice, gives \(\operatorname{rank}(B_{\infty,p})=|\mathcal S|\). The limiting rank-\(h\) operator error is \(\sigma_{h+1}(B_{\infty,p})/\sigma_1(B_{\infty,p})>0\) whenever \(h<|\mathcal S|\).
\end{proof}

\subsection{Absolute approximation risk}

Write $\widetilde B_{H,p}=\lambda_1^{-(H-1)}B_{H,p}$ and assume $\Gamma_{L,p}\preceq M_\Gamma I$. The operator-norm bound from Theorem~\ref{thm:floor} then implies
\begin{equation}
 \|\widetilde B_{H,p}-B_{\infty,p}\|_{\Gamma_{L,p}}
 \le
 \delta_{H,p},\qquad
 \delta_{H,p}=\sqrt{M_\Gamma\min\{N,p\}}\,C_pq_A^{H-1}.
\end{equation}
The rest of the argument converts this matrix bound into a bound on the best rank-$h$ approximation error.

Let \(\mathcal R_h\) be the closed cone of matrices of rank at most \(h\), and let \(d_\Gamma(B,\mathcal R_h)=\inf_{M\in\mathcal R_h}\|B-M\|_{\Gamma_{L,p}}\). Distance to any nonempty set is 1-Lipschitz in the ambient seminorm, hence
\begin{equation}
\left|d_\Gamma(\widetilde B_{H,p},\mathcal R_h)
-d_\Gamma(B_{\infty,p},\mathcal R_h)\right|
\le\|\widetilde B_{H,p}-B_{\infty,p}\|_{\Gamma_{L,p}}.
\end{equation}
The covariance upper bound, the inequality \(\|D\|_{\mathrm{F}}\le\sqrt{\min\{N,p\}}\|D\|_{\mathrm{op}}\), and Theorem~\ref{thm:floor} bound the right-hand side by \(\delta_{H,p}\). This proves the root-risk bound. Since \(|a^2-b^2|=|a-b|(a+b)\) and \(a\le b+|a-b|\), the squared-risk bound follows. Finally, \(\mathcal R_h\) is a cone, so for every nonzero scalar \(c\), \(d_\Gamma(cB,\mathcal R_h)=|c|d_\Gamma(B,\mathcal R_h)\). Taking \(c=\lambda_1^{H-1}\) proves the displayed absolute expansion.

\subsection{Two-mode closed form}

For $r=2$, write $B_{H,p}=U_pD_HV_p^\top$, where $U_p=[u_{N,p}(\varphi_1),u_{N,p}(\varphi_2)]$, $V_p=[v_p(\varphi_1),v_p(\varphi_2)]$, and $D_H=\operatorname{diag}(d_1(H),d_2(H))$. Let $G_U=U_p^*U_p$ and $G_V=V_p^*V_p$. The nonzero squared singular values are the eigenvalues of
\begin{equation}
G_U^{1/2}D_H\overline{G_V}D_H^*G_U^{1/2}.
\end{equation}
Define
\begin{equation}
T_H=\operatorname{tr}(G_UD_H\overline{G_V}D_H^*),
\qquad
\Delta_H=\det(G_U)\det(G_V)|d_1(H)d_2(H)|^2.
\end{equation}
Then
\begin{equation}
\sigma_{1,2}^2(B_{H,p})=
\frac{T_H\pm\sqrt{T_H^2-4\Delta_H}}{2}.
\end{equation}
Here \(\overline{G_V}=V_p^\top\overline{V_p}\). For real modes it equals \(G_V\), recovering the simpler real formula. The Gram entries are finite geometric sums. Replacing $D_H$ by $D_\infty=\operatorname{diag}(g_{11},g_{12})$ gives the exact two-mode floor. For rank-one approximation with \(N,p\ge2\), the two-mode floor is positive exactly when the limiting matrix has rank two. Under two active modes this is equivalent to \(g_{11}g_{12}\ne0\), \(\varphi_1\ne\varphi_2\), and \(\varphi_1^p\ne\varphi_2^p\).

\subsection{Filtering patterns shared across horizons}
\label{sec:multi-horizon}
For a positive integer $T$ of consecutive horizons, define
\begin{equation}
 \mathcal B_{T,p}[q,j,s]=CA^{q-1}F^{jp+s}K,\quad 1\le q\le T.
 \label{eq:tensor}
\end{equation}
The factorization gives
\begin{equation}
 \mathcal B_{T,p}[q,j,s]=\sum_{k=1}^r d_k(q)
 u_{N,p}(\varphi_k)_jv_p(\varphi_k)_s.
 \label{eq:cp}
\end{equation}
The same filtering patterns can therefore be reused across horizons, with different weights $d_k(q)$. Formally, the tensor's canonical polyadic rank, the minimum number of separable three-factor terms, is at most $r$ over $\mathbb C$. The real bound requires real factors. Reusing patterns suggests a joint model, but does not prove that learning horizons together reduces error. Any weights used to combine horizon-specific risks should be fixed before testing.

\begin{proof}[Derivation of \eqref{eq:cp}]
Using the spectral decompositions of \(A\) and \(F\),
\begin{equation}
CA^{q-1}F^{jp+s}K
=\sum_{a=1}^r\sum_{k=1}^r
g_{ak}\lambda_a^{q-1}(\varphi_k^p)^j\varphi_k^s
=\sum_{k=1}^r d_k(q)
u_{N,p}(\varphi_k)_jv_p(\varphi_k)_s.
\end{equation}
For each \(k\), the dependence on \(q\), \(j\), and \(s\) separates into three factors. The tensor is therefore a sum of at most \(r\) rank-one CP components over \(\mathbb C\). If all factors are real, this is also a real CP-rank bound. With nonreal conjugate modes, no unqualified equality between real and complex CP rank is asserted.
\end{proof}

\section{FINITE-WINDOW PREDICTION AND APPROXIMATION}
\label{sec:estimation}

This section supplies the population bridge needed for patch selection. It first records how the covariance geometry changes low-rank approximation, then bounds the effect of replacing the infinite past by a finite observation window. The final result expresses that bound in the notation of Section~\ref{sec:selection}.

\subsection{Covariance-weighted approximation}

The coefficient matrix is judged in prediction norm, not in ordinary Frobenius norm. The next proposition explains when the familiar singular-value picture remains exact and, more generally, how covariance eigenvalue bounds translate between the two norms.

\begin{proposition}[Approximation risk under structured designs]
If $\Gamma_{L,p}=\Sigma_p\otimes\Sigma_N$ with positive-definite factors, then $\mathcal{E}_h^\Gamma(B)$ equals the squared singular-value tail of $\Sigma_N^{1/2}B\Sigma_p^{1/2}$. In particular, under isotropy it equals $\sum_{j>h}\sigma_j^2(B)$. More generally, if $m_\Gamma I\preceq\Gamma_{L,p}\preceq M_\Gamma I$, then
\begin{equation}
 m_\Gamma\sum_{j>h}\sigma_j^2(B)
 \le \mathcal{E}_h^\Gamma(B)
 \le M_\Gamma\sum_{j>h}\sigma_j^2(B).
 \label{eq:sandwich}
\end{equation}
\end{proposition}

\begin{proof}

Under column-major vectorization and a Kronecker covariance,
\begin{equation}
\operatorname{vec}(E)^\top(\Sigma_p\otimes\Sigma_N)\operatorname{vec}(E)
=\|\Sigma_N^{1/2}E\Sigma_p^{1/2}\|_{\mathrm{F}}^2.
\end{equation}
Because the two covariance factors are positive definite, the map \(M\mapsto\Sigma_N^{1/2}M\Sigma_p^{1/2}\) is bijective and preserves rank. The weighted rank-\(h\) problem is therefore exactly the ordinary Frobenius rank-\(h\) approximation of \(\Sigma_N^{1/2}B\Sigma_p^{1/2}\), proving the singular-value-tail formula by Eckart--Young.

For any error matrix $E$,
\begin{equation}
m_\Gamma\|E\|_{\mathrm{F}}^2
\le\operatorname{vec}(E)^\top\Gamma_{L,p}\operatorname{vec}(E)
\le M_\Gamma\|E\|_{\mathrm{F}}^2.
\end{equation}
Taking the infimum over $E=B-M$ with $\operatorname{rank}(M)\le h$ proves the lower bound in \eqref{eq:sandwich}. Evaluating the upper quadratic form at the ordinary truncated-SVD optimizer gives the upper bound. Under isotropy both inequalities are equalities.
\end{proof}

\subsection{The finite-window bridge}

Truncating an infinite-past forecast leaves its retained weights unchanged. A predictor restricted to the window can do better by refitting those weights. Define $b^\circ_{H,L}=\arg\min_b\mathbb E(X_{t+H}-b^\top x_{t,L})^2$ and let $B^\circ_{H,L,p}$ be its patch reshape. The superscript $\circ$ distinguishes this finite-window target from $B_{H,p}$.

Let $\mathsf M_X$ be the essential upper bound of the observation spectral density, normalized so that $\operatorname{Cov}(X_{t+k},X_t)=(2\pi)^{-1}\int_{-\pi}^{\pi}e^{\mathrm i k\omega}f_X(\omega)\,d\omega$. Stability and $R>0$ imply positive lower and finite upper spectral bounds, hence a unique finite-window target. Choose $K_H<\infty$ and $\rho_F < \bar \rho_F<1$ with $|\beta_H(\ell)|\le K_H\bar\rho_F^\ell$, and set $D_{H,L}=\mathsf M_XK_H^2\bar\rho_F^{2L}/(1-\bar\rho_F^2)$.

\begin{theorem}[Finite-window approximation bound]
\label{thm:bridge}
Uniformly over divisors $p$ of $L$,
\begin{align}
\|B^\circ_{H,L,p}-B_{H,p}\|_{\Gamma_{L,p}}^2&\le D_{H,L}, \label{eq:bridge1} \\
\left|\sqrt{\mathcal{E}_h^\Gamma(B^\circ_{H,L,p})}
 -\sqrt{\mathcal{E}_h^\Gamma(B_{H,p})}\right|&\le\sqrt{D_{H,L}}. \label{eq:bridge2}
\end{align}
If $\Gamma_{L,p}\succeq m_\Gamma I$, the squared Frobenius difference is at most $D_{H,L}/m_\Gamma$.
\end{theorem}

\begin{proof}
The omitted old observations contribute a tail to the infinite-past forecast. Refitting the retained window can recover only the part of that tail predictable from the retained observations. Orthogonal projection cannot increase its mean squared size. This is why the same tail bound controls the change in coefficients measured by prediction error.

Let \(b_{H,L}=(\beta_H(0),\ldots,\beta_H(L-1))^\top\), ordered consistently with \(x_{t,L}=(X_t,\ldots,X_{t-L+1})^\top\), and write the full-past forecast as
\begin{equation}
\Pi_{t,H}=b_{H,L}^\top x_{t,L}+T_{t,H,L},
\qquad
T_{t,H,L}=\sum_{\ell\ge L}\beta_H(\ell)X_{t-\ell}.
\end{equation}
Let $\mathsf P_L$ be orthogonal projection in $L_2$ onto the span of $x_{t,L}$. Since the full-past forecast error is orthogonal to the observed past,
\begin{equation}
(b_{H,L}^\circ)^\top x_{t,L}=\mathsf P_L\Pi_{t,H}.
\end{equation}
Projection contraction and the spectral-density upper bound give
\begin{equation}
\mathbb E\{\Pi_{t,H}-\mathsf P_L\Pi_{t,H}\}^2
\le\mathbb ET_{t,H,L}^2
\le \mathsf M_X\sum_{\ell\ge L}\beta_H(\ell)^2
\le D_{H,L}.
\end{equation}
The middle inequality follows directly from the stated Fourier convention and Parseval's identity: for every square-summable scalar filter \(a\), \(\operatorname{Var}(\sum_\ell a_\ell X_{t-\ell}) \le \mathsf M_X\sum_\ell|a_\ell|^2\). Moreover,
\begin{equation}
(b_{H,L}^\circ-b_{H,L})^\top x_{t,L}
=\mathsf P_LT_{t,H,L},
\end{equation}
and projection contraction gives
\begin{equation}
\mathbb E(\mathsf P_LT_{t,H,L})^2
\le \mathbb ET_{t,H,L}^2\le D_{H,L}.
\end{equation}
This proves \eqref{eq:bridge1} after the fixed permutation from lag order to patch order. Positive definiteness of \(\Gamma_{L,p}\) makes the finite-window coefficient unique. The square root of approximation risk is distance to the closed rank-at-most-$h$ set in the $\Gamma_{L,p}$ norm, hence is 1-Lipschitz. This proves \eqref{eq:bridge2}. The spectral lower bound gives the Frobenius conclusion.
\end{proof}

Because the bound decreases geometrically with $L$, the spectrum of the truncated coefficient matrix describes the finite-window approximation problem increasingly well. Exact low rank of one target, however, does not imply exact low rank of the other.

\subsection{Transfer to the patch-selection approximation error}
\label{app:population-transfer}

\begin{theorem}[Finite-window approximation]
\label{thm:patch-population-transfer}
Under the state-space and stable-filter assumptions of Section~\ref{sec:model}, for every divisor $p$ of $L$,
\begin{align}
\left|\sqrt{A_m(p)}-\sqrt{\mathcal{E}_m^\Gamma(B_{H,p})}\right|
 &\le \|B_p^\circ-B_{H,p}\|_\Gamma, \label{eq:patch-population-transfer} \\
0\le A_m(p)&\le\|B_p^\circ-B_{H,p}\|_\Gamma^2
 \quad\text{if }m\ge r. \label{eq:patch-finite-window-approximation}
\end{align}
The squared discrepancy $\|B_p^\circ-B_{H,p}\|_\Gamma^2$ admits a geometrically decreasing bound in $L$, uniformly over patch lengths, as quantified in Theorem~\ref{thm:bridge}.
\end{theorem}

\begin{proof}
For any $B\in\mathcal{M}_m(p)$, the triangle inequality gives
\begin{equation}
 \sqrt{A_m(p)}
 \le \|B_p^\circ-B\|_\Gamma
 \le \|B_p^\circ-B_{H,p}\|_\Gamma
     +\|B_{H,p}-B\|_\Gamma.
\end{equation}
Taking the infimum over $B$, and then interchanging $B_p^\circ$ and $B_{H,p}$ in the same argument, proves \eqref{eq:patch-population-transfer}.

Theorem~\ref{thm:bridge} gives the stated geometric bound on $\|B_p^\circ-B_{H,p}\|_\Gamma^2$, uniformly in $p$. Finally, Theorem~\ref{thm:factor} gives $\operatorname{rank}(B_{H,p})\le r$. If $m\ge r$, $B_{H,p}$ is feasible in the minimization defining $A_m(p)$, which yields \eqref{eq:patch-finite-window-approximation}.
\end{proof}

\section{ESTIMATION FROM ONE DEPENDENT TRAJECTORY}
\label{app:dependent-rate}

The estimation argument has three stages. A deterministic oracle inequality reduces prediction error to curvature, score, and comparator bounds. Gaussian concentration then establishes these bounds simultaneously over the candidate patches, ranks, and horizons. Combining the two gives the uniform rate and the finite-sample theorem in the main paper. Universal constants are denoted by $c,C$ and may change from line to line. Readers interested in the final bounds can go directly to Corollary~\ref{cor:uniform-dependent-rate} and the proof of Theorem~\ref{thm:patch-estimation} below.

\subsection{Deterministic oracle inequality}
\label{app:oracle}

The proof compares the fitted coefficients with the best allowed population coefficients. It bounds how much the training loss can favor an incorrect coefficient difference because of noise. The three assumed events are proved for the dependent Gaussian data in the subsections below.

Fix a candidate patch length $p$, rank bound $h$, and horizon $H$. Let $\mathcal I_n=\{1,\ldots,n\}$ and define
\begin{equation}
 Q_p(D)=\mathbb E\langle D,Z_{t,p}\rangle_{\mathrm{F}}^2,
 \qquad
 \widehat Q_p(D)=\frac1n\sum_{t\in\mathcal I_n}
 \langle D,Z_{t,p}\rangle_{\mathrm{F}}^2.
\end{equation}
Write $\widehat B_{p,h}=\widehat B_{p,h,H}$ for a global rank-constrained empirical least-squares minimizer and $u_{t,H,L}=X_{t+H}-\langle B^\circ_{H,L,p},Z_{t,p}\rangle_{\mathrm{F}}$ for its population residual. The regression score matrix is $G_p=n^{-1}\sum_{t\in\mathcal I_n}u_{t,H,L}Z_{t,p}$. Let $B_h^\circ$ be a population rank-$h$ optimizer and $E=B_h^\circ-B^\circ_{H,L,p}$. Assume constants $\kappa_p,M_p>0$ such that, for every rank-at-most-$2h$ coefficient difference $D$, $\widehat Q_p(D)\ge\kappa_p\|D\|_{\mathrm{F}}^2$ and $Q_p(D)\le M_p\|D\|_{\mathrm{F}}^2$. Also assume $\widehat Q_p(E)\le2Q_p(E)$. These are the restricted-curvature, population-upper-bound, and oracle-residual conditions used below.

\begin{theorem}[Oracle inequality on dependent-data events]
\label{thm:oracle}
On the restricted-curvature, population-upper-bound, and oracle-residual events stated above and verified in Appendix~\ref{app:dependent-rate},
\begin{align}
Q_p(\widehat B_{p,h}-B^\circ_{H,L,p})
&\le \left(2+\frac{32M_p}{\kappa_p}\right)
\mathcal{E}_h^\Gamma(B^\circ_{H,L,p}) \nonumber\\
&\quad+\frac{64M_ph}{\kappa_p^2}\|G_p\|_{\mathrm{op}}^2. \label{eq:oracle}
\end{align}
\end{theorem}

\begin{proof}
Let $B_h^\circ$ be a population rank-$h$ optimizer, set $E=B_h^\circ-B^\circ_{H,L,p}$ and $\Delta=\widehat B_{p,h}-B_h^\circ$. Then $\operatorname{rank}(\Delta)\le2h$. The empirical basic inequality is
\begin{equation}
\widehat Q_p(E+\Delta)
\le \widehat Q_p(E)+2\langle G_p,\Delta\rangle_{\mathrm{F}}.
\end{equation}
Using $\|a+b\|^2\ge\frac12\|b\|^2-\|a\|^2$ in the empirical prediction norm,
\begin{equation}
\widehat Q_p(\Delta)
\le4\widehat Q_p(E)+4|\langle G_p,\Delta\rangle_{\mathrm{F}}|.
\end{equation}
Nuclear/operator duality and $\operatorname{rank}(\Delta)\le2h$ imply
\begin{equation}
|\langle G_p,\Delta\rangle_{\mathrm{F}}|
\le\|G_p\|_{\mathrm{op}}\|\Delta\|_*
\le\sqrt{2h}\|G_p\|_{\mathrm{op}}\|\Delta\|_{\mathrm{F}}.
\end{equation}
Restricted curvature, the assumed oracle-residual concentration $\widehat Q_p(E)\le2Q_p(E)$, and Young's inequality give
\begin{equation}
\|\Delta\|_{\mathrm{F}}^2
\le\frac{16}{\kappa_p}Q_p(E)
+\frac{32h}{\kappa_p^2}\|G_p\|_{\mathrm{op}}^2.
\end{equation}
Finally,
\begin{equation}
Q_p(E+\Delta)
\le2Q_p(E)+2Q_p(\Delta)
\le\left(2+\frac{32M_p}{\kappa_p}\right)Q_p(E)
+\frac{64M_ph}{\kappa_p^2}\|G_p\|_{\mathrm{op}}^2,
\end{equation}
and $Q_p(E)=\mathcal{E}_h^\Gamma(B^\circ_{H,L,p})$.

For completeness, \(\Gamma_{L,p}\succ0\) makes the population objective coercive on the closed rank-at-most-\(h\) variety, so \(B_h^\circ\) exists. On the stated restricted-curvature event, \(\widehat Q_p(B)\ge\kappa_p\|B\|_{\mathrm{F}}^2\) for every rank-at-most-\(h\) matrix \(B\). Hence the empirical loss is also coercive on that closed variety and a global empirical minimizer exists. The result concerns such a global minimizer and does not assert that a particular nonconvex algorithm finds it.
\end{proof}

\subsection{Gaussian model and spectral bounds}

Assume the scalar stationary process is generated by
\begin{equation}
S_{t+1}=AS_t+\eta_{t+1},\qquad X_t=CS_t+\xi_t,
\end{equation}
where \(\rho_A<1\), \(\eta_t\stackrel{\mathrm{iid}}\sim N(0,Q)\), \(\xi_t\stackrel{\mathrm{iid}}\sim N(0,R)\), \(R>0\), and the two noise sequences are mutually independent.  With the convention
\begin{equation}
\gamma_X(k)=\frac1{2\pi}\int_{-\pi}^{\pi}
e^{\mathrm i k\omega}f_X(\omega)\,d\omega,
\end{equation}
the spectral density is
\begin{equation}
f_X(\omega)=
C(I-Ae^{-\mathrm i\omega})^{-1}Q
(I-A^\top e^{\mathrm i\omega})^{-1}C^\top+R.
\end{equation}
Define
\begin{equation}
\mathsf m_X=\operatorname*{ess\,inf}_{\omega}f_X(\omega),\qquad
\mathsf M_X=\operatorname*{ess\,sup}_{\omega}f_X(\omega),\qquad
\chi_X=\mathsf M_X/\mathsf m_X.
\label{eq:spectral-condition}
\end{equation}
These quantities are finite and strictly positive.  Indeed, with
\begin{equation}
\mathcal R_A=\sup_{|z|=1}\|(I-Az)^{-1}\|_{\mathrm{op}},
\end{equation}
stability and \(R>0\) give the explicit bounds
\begin{equation}
R\le \mathsf m_X\le\mathsf M_X
\le R+\|C\|_2^2\|Q\|_{\mathrm{op}}\mathcal R_A^2.
\label{eq:ssm-spectral-bound}
\end{equation}
Thus \(\chi_X\), unlike \(\rho_A\) alone, records both persistence and nonnormal resolvent amplification.

\subsection{Lag filters and candidate classes}

Fix a context length \(L\).  For every patch \(p\mid L\), put \(N_p=L/p\), and let \(Z_{t,p}\in\mathbb R^{N_p\times p}\) be the fixed permutation/reshape of \(x_{t,L}=(X_t,\ldots,X_{t-L+1})^\top\).  For any matrix \(D\), let \(\operatorname{lag}_p(D)\in\mathbb R^L\) be its coefficient vector in lag order and define
\begin{equation}
a_{p,D}(\omega)=\sum_{\ell=0}^{L-1}\operatorname{lag}_p(D)_\ell e^{-\mathrm i\ell\omega}.
\end{equation}
Because patching is a permutation,
\begin{equation}
\|\operatorname{lag}_p(D)\|_2=\|D\|_{\mathrm{F}},
\qquad
\sup_\omega|a_{p,D}(\omega)|^2
\le L\|D\|_{\mathrm{F}}^2.                              \label{eq:fourier-envelope}
\end{equation}
Consequently,
\begin{equation}
\mathsf m_X\|D\|_{\mathrm{F}}^2
\le Q_p(D):=\mathbb E\langle D,Z_{t,p}\rangle_{\mathrm{F}}^2
\le\mathsf M_X\|D\|_{\mathrm{F}}^2.                    \label{eq:population-curvature}
\end{equation}
For the consecutive training set \(\mathcal I_n=\{1,\ldots,n\}\), define
\begin{equation}
\widehat Q_p(D)=\frac1n\sum_{t=1}^n
\langle D,Z_{t,p}\rangle_{\mathrm{F}}^2.
\label{eq:empirical-quadratic}
\end{equation}

Let \(\mathcal P_L\) be a nonempty finite set of admissible divisors, let \(\mathcal H\subset\{1,2,\ldots\}\) be a nonempty finite set of horizons, and, for every \(p\), let \(\mathcal R_p\) be a nonempty finite set of positive candidate ranks.  Assume \(0<\delta<1\).  Put
\begin{equation}
\mathfrak K=|\mathcal H|\sum_{p\in\mathcal P_L}|\mathcal R_p|,
\qquad
\ell_\delta=\log\frac{12\max\{1,\mathfrak K\}}{\delta}.
\end{equation}
There are \(\sum_p|\mathcal R_p|\) curvature classes, \(2|\mathcal P_L||\mathcal H|\) signed score-polarization classes, and \(\mathfrak K\) fixed-oracle classes.  Nonemptiness implies that their total is at most \(4\mathfrak K\). This is the union-class count used with \(\ell_\delta\). For \((p,h)\), define
\begin{align}
r_{p,h}&=\min\{2h,N_p,p\},\qquad
s_{p,h}=r_{p,h}(N_p+p+1), \label{eq:entropy-size} \\
k_{p,h}&=s_{p,h}+\ell_\delta,\qquad
q_p=N_p+p+1+\ell_\delta. \label{eq:complexity-sizes}
\end{align}

\subsection{Forecast residuals and spectral scales}

For each \(H\), let \(b^\circ_{H,L}\) be the population least-squares coefficient for predicting \(X_{t+H}\) from \(x_{t,L}\).  Its residual
\begin{equation}
u_{t,H,L}=X_{t+H}-(b^\circ_{H,L})^\top x_{t,L}
\end{equation}
is jointly Gaussian with the lag vector and is independent of that lag vector at the same time because population orthogonality and joint Gaussianity imply independence.  Define the residual transfer polynomial, variance, and spectral maximum by
\begin{align}
q_{H,L}(\omega)&=e^{\mathrm iH\omega}
-\sum_{\ell=0}^{L-1}b^\circ_{H,L}(\ell)e^{-\mathrm i\ell\omega}, \\
\sigma_{u,H,L}^2&=\frac1{2\pi}\int_{-\pi}^{\pi}
f_X(\omega)|q_{H,L}(\omega)|^2\,d\omega, \\
\mathsf M_{u,H,L}&=\operatorname*{ess\,sup}_{\omega}
f_X(\omega)|q_{H,L}(\omega)|^2, \label{eq:residual-spectrum} \\
\mathsf S_{H,L}&=\sigma_{u,H,L}^2+\mathsf M_X, \nonumber\\
\Lambda_{H,L}&=\mathsf M_{u,H,L}+L\mathsf M_X. \label{eq:score-stability}
\end{align}
All of these are finite deterministic functionals of the SSM and the finite-window projection.

The one-sided sufficient restricted-curvature scale used below is
\begin{equation}
n_{\mathrm{curv}}^{\mathrm{suff}}(L)=\frac{n}{\chi_X^2L}.
\label{eq:explicit-neff}
\end{equation}
It is only a conservative sufficient scale for the curvature proof.  The factor \(L\) is the exact Fourier-envelope penalty in \eqref{eq:fourier-envelope}, not an independence-blocking assumption.

\subsection{Gaussian concentration tools}

For a metric space $(\mathcal T,d)$, define $\gamma_2(\mathcal T,d)=\inf_{(\mathcal T_j)}\sup_{t\in\mathcal T} \sum_{j\ge0}2^{j/2}d(t,\mathcal T_j)$, where $d(t,\mathcal T_j)=\inf_{s\in\mathcal T_j}d(t,s)$, $|\mathcal T_0|=1$, and $|\mathcal T_j|\le2^{2^j}$. This is the generic-chaining functional used in the following Gaussian chaos bound.  It is the Gaussian specialization obtained from Theorem~3.5 and Proposition~2.6 of \citet{krahmer2014chaos}.

\begin{lemma}[Gaussian chaos supremum]
\label{lem:gaussian-chaos}
Let \(g\sim N(0,I_m)\) and let \(\mathcal A\) be a bounded set of deterministic real matrices with a common number of columns.  Define
\begin{equation}
d_F=\sup_{A\in\mathcal A}\|A\|_{\mathrm{F}},
\quad d_{\mathrm{op}}=\sup_{A\in\mathcal A}\|A\|_{\mathrm{op}},
\quad \gamma=\gamma_2(\mathcal A,\|\cdot\|_{\mathrm{op}}).
\end{equation}
There are universal constants \(C,c>0\) such that, for every \(v\ge1\), with probability at least \(1-2e^{-cv^2}\),
\begin{align}
\sup_{A\in\mathcal A}
\left|\|Ag\|_2^2-\mathbb E\|Ag\|_2^2\right|
\le C\{&\gamma^2+d_F\gamma+d_Fd_{\mathrm{op}} \nonumber\\
&+v d_{\mathrm{op}}(\gamma+d_F)+v^2d_{\mathrm{op}}^2\}. \label{eq:kmr-bound}
\end{align}
\end{lemma}

The assumptions of this imported result are met below because the entire finite observation vector is Gaussian and can be written as a deterministic covariance square root times a standard Gaussian vector. All indexing matrices are deterministic population objects.  The proof also uses the fixed-matrix Hanson--Wright inequality, Theorem~1.1 of \citet{rudelson2013hanson}.

Let $\mathcal N(\mathcal T,d,\epsilon)$ denote the smallest number of radius-$\epsilon$ balls in metric $d$ covering $\mathcal T$. The rank-class entropy bound is
\begin{equation}
\log\mathcal N\!\left(
\{D:\operatorname{rank}(D)\le r,\ \|D\|_{\mathrm{F}}=1\},
\|\cdot\|_{\mathrm{F}},\epsilon\right)
\le r(N_p+p+1)\log(9/\epsilon),                    \label{eq:rank-cover}
\end{equation}
for \(0<\epsilon<1\).  This is Lemma~3.1 of \citet{candes2011tight} and also follows directly by covering the two Stiefel factors and singular values in an SVD.  Dudley's entropy integral applied to \eqref{eq:rank-cover} is finite and contributes the square root of the exponent in \eqref{eq:entropy-size}.

\subsection{Simultaneous dependent-data events}

Use the consecutive training targets \(\mathcal I_n=\{1,\ldots,n\}\).  All variables needed by every horizon and lag lie in the single Gaussian vector
\begin{equation}
\mathbf X=(X_{1-L+1},\ldots,X_{n+\max\mathcal H})^\top.
\end{equation}
Thus the following result accounts for the full overlap among the training examples, whose context-to-target spans have lengths \(L+H\).

\begin{theorem}[Uniform dependent-data events]
\label{thm:uniform-dependent-events}
There is a universal constant \(C>0\) such that, with probability at least \(1-\delta\), the following statements hold simultaneously for every \(p\in\mathcal P_L\), \(h\in\mathcal R_p\), and \(H\in\mathcal H\).

First, for every \(D\in\mathbb R^{N_p\times p}\) with \(\operatorname{rank}(D)\le r_{p,h}\),
\begin{equation}
|\widehat Q_p(D)-Q_p(D)|
\le C\mathsf M_X
\left\{\sqrt{\frac{Lk_{p,h}}n}
+\frac{Lk_{p,h}}n\right\}\|D\|_{\mathrm{F}}^2.
\label{eq:uniform-rsc-deviation}
\end{equation}

Second, for
\begin{align}
G_{p,H}&=\frac1n\sum_{t=1}^n u_{t,H,L}Z_{t,p},\\
\|G_{p,H}\|_{\mathrm{op}}
&\le C\left\{
\sqrt{\frac{\mathsf S_{H,L}\Lambda_{H,L}q_p}{n}}
+\frac{\Lambda_{H,L}q_p}{n}
\right\}. \label{eq:uniform-score}
\end{align}

Third, let \(B^\circ_{p,H}=B^\circ_{H,L,p}\) be the patch reshape of \(b^\circ_{H,L}\), let
\begin{equation}
B^\circ_{p,h,H}\in\arg\min_{\operatorname{rank}(B)\le h}
Q_p(B-B^\circ_{p,H}),
\qquad
E_{p,h,H}=B^\circ_{p,h,H}-B^\circ_{p,H}.
\end{equation}
Then
\begin{equation}
|\widehat Q_p(E_{p,h,H})-Q_p(E_{p,h,H})|
\le C Q_p(E_{p,h,H})
\left\{\sqrt{\frac{\chi_XL\ell_\delta}{n}}
+\frac{\chi_XL\ell_\delta}{n}\right\}.
\label{eq:uniform-oracle-residual}
\end{equation}
If \(E_{p,h,H}=0\), the left side is identically zero.
\end{theorem}

\begin{proof}
\textit{Step 1: restricted curvature.} Represent \(\mathbf X=\Sigma_{\mathbf X}^{1/2}g\) with \(g\sim N(0,I)\).  For a lag coefficient \(d=\operatorname{lag}_p(D)\), let \(T_d\mathbf X=((d^\top x_{t,L}))_{t=1}^n\) and define \(A_D=n^{-1/2}T_d\Sigma_{\mathbf X}^{1/2}\).  Stationarity and the spectral integral give
\begin{align}
\|A_D\|_{\mathrm{F}}^2&=Q_p(D)\le\mathsf M_X\|D\|_{\mathrm{F}}^2, \nonumber\\
\|A_D-A_{D'}\|_{\mathrm{op}}
&\le\sqrt{\frac{\mathsf M_XL}{n}}\|D-D'\|_{\mathrm{F}}. \label{eq:convolution-map-bounds}
\end{align}
The second line follows because the covariance matrix of a length-\(n\) sample from a process with spectral density \(f_X|a_{p,D-D'}|^2\) has operator norm at most its spectral supremum, followed by \eqref{eq:fourier-envelope}.

Apply Lemma~\ref{lem:gaussian-chaos} to \(\mathcal A_{p,h}=\{A_D:\operatorname{rank}(D)\le r_{p,h},\|D\|_{\mathrm{F}}=1\}\). Equations \eqref{eq:rank-cover}--\eqref{eq:convolution-map-bounds} and Dudley's integral give
\begin{align}
d_F&\le\sqrt{\mathsf M_X},\qquad
d_{\mathrm{op}}\le\sqrt{\mathsf M_XL/n}, \nonumber\\
\gamma_2(\mathcal A_{p,h},\|\cdot\|_{\mathrm{op}})
&\le C\sqrt{\frac{\mathsf M_XL}{n}}\sqrt{s_{p,h}}.
\end{align}
With \(v^2\) proportional to \(\ell_\delta\), the six terms in \eqref{eq:kmr-bound} are bounded, respectively, by universal multiples of
\begin{align}
&\frac{\mathsf M_XLs_{p,h}}n,\qquad
\mathsf M_X\sqrt{\frac{Ls_{p,h}}n},\qquad
\mathsf M_X\sqrt{\frac Ln}, \nonumber\\
&\frac{\mathsf M_XL\sqrt{s_{p,h}\ell_\delta}}n,\qquad
\mathsf M_X\sqrt{\frac{L\ell_\delta}n},\qquad
\frac{\mathsf M_XL\ell_\delta}n.
\end{align}
Using \(\sqrt{s\ell}\le(s+\ell)/2\) shows term by term that their sum is bounded by the right side of \eqref{eq:uniform-rsc-deviation} for \(\|D\|_{\mathrm{F}}=1\).  Both quadratic forms are homogeneous of degree two, so rescaling gives the claim for every nonzero \(D\). It is trivial for \(D=0\).  The stated union-class count then proves the simultaneous curvature event.

\textit{Step 2: score concentration.} For the score, let \(U_H\mathbf X=(u_{t,H,L})_{t=1}^n\) and put \(R_H=n^{-1/2}U_H\Sigma_{\mathbf X}^{1/2}\).  The residual spectral density and contemporaneous projection orthogonality give
\begin{align}
\|R_H\|_{\mathrm{F}}^2&=\sigma_{u,H,L}^2,\qquad
\|R_H\|_{\mathrm{op}}^2\le\mathsf M_{u,H,L}/n, \nonumber\\
\operatorname{tr}(R_HA_D^\top)
&=\mathbb E\langle R_Hg,A_Dg\rangle \nonumber\\
&=\frac1n\sum_{t=1}^n
\mathbb E\{u_{t,H,L}\langle D,Z_{t,p}\rangle_{\mathrm{F}}\} \nonumber\\
&=0\qquad\text{for every }D.
\end{align}
The last equality is the trace orthogonality needed for centered polarization. It follows because \(b^\circ_{H,L}\) is the population projection onto the full \(L\)-lag span. Operator/nuclear duality gives
\begin{equation}
\|G_{p,H}\|_{\mathrm{op}}
=\sup_{\operatorname{rank}(D)=1,\|D\|_{\mathrm{F}}=1}
|\langle R_Hg,A_Dg\rangle|.
\end{equation}
For every such \(D\), polarization yields
\begin{equation}
\langle R_Hg,A_Dg\rangle
=\frac14\{\|(R_H+A_D)g\|_2^2-\|(R_H-A_D)g\|_2^2\}.
\end{equation}
Apply Lemma~\ref{lem:gaussian-chaos} to the two matrix classes \(\{R_H\pm A_D:\operatorname{rank}(D)=1,\|D\|_{\mathrm{F}}=1\}\). Their parameters satisfy
\begin{align}
d_F^2&\le\mathsf S_{H,L},\qquad
d_{\mathrm{op}}^2\le 2\Lambda_{H,L}/n, \nonumber\\
\gamma_2&\le C\sqrt{\frac{\mathsf M_XL}{n}}\sqrt{N_p+p+1}.
\end{align}
Write \(d_p^0=N_p+p+1\), so \(q_p=d_p^0+\ell_\delta\).  Term by term, the six quantities in \eqref{eq:kmr-bound} satisfy
\begin{align}
\gamma_2^2&\le C\Lambda_{H,L}q_p/n, \nonumber\\
d_F\gamma_2&\le C\sqrt{\mathsf S_{H,L}\Lambda_{H,L}q_p/n}, \nonumber\\
d_Fd_{\mathrm{op}}&\le C\sqrt{\mathsf S_{H,L}\Lambda_{H,L}q_p/n}, \nonumber\\
v d_{\mathrm{op}}\gamma_2
&\le C\Lambda_{H,L}\sqrt{d_p^0\ell_\delta}/n
\le C\Lambda_{H,L}q_p/n, \nonumber\\
v d_{\mathrm{op}}d_F
&\le C\sqrt{\mathsf S_{H,L}\Lambda_{H,L}q_p/n}, \nonumber\\
v^2d_{\mathrm{op}}^2&\le C\Lambda_{H,L}q_p/n.
\end{align}
Here we used \(\mathsf M_XL\le\Lambda_{H,L}\), \(\sqrt{d_p^0\ell_\delta}\le q_p/2\), and \(q_p\ge1\). This proves \eqref{eq:uniform-score}. The two signs and the stated union-class count give simultaneity over \((p,H)\).

\textit{Step 3: the fixed population comparator.} Finally, for a fixed nonzero \(E=E_{p,h,H}\), the same convolution representation has
\begin{equation}
\|A_E\|_{\mathrm{F}}^2=Q_p(E),
\qquad
\frac{\|A_E\|_{\mathrm{op}}^2}{Q_p(E)}
\le\frac{\chi_XL}{n}.
\end{equation}
The fixed-matrix Hanson--Wright inequality applied to \(\|A_Eg\|_2^2\), followed by a union bound over the finite triples, proves \eqref{eq:uniform-oracle-residual}.
\end{proof}

\subsection{Uniform rank-constrained regression rate}

For every triple, let a global empirical rank-constrained estimator satisfy
\begin{equation}
\widehat B_{p,h,H}\in\arg\min_{\operatorname{rank}(B)\le h}
\frac1n\sum_{t=1}^n
\{X_{t+H}-\langle B,Z_{t,p}\rangle_{\mathrm{F}}\}^2,
\end{equation}
and define the approximation error
\begin{equation}
\mathcal A_{p,h,H}=Q_p(B^\circ_{p,h,H}-B^\circ_{p,H}).
\end{equation}

\begin{corollary}[Uniform one-trajectory oracle rate]
\label{cor:uniform-dependent-rate}
\label{cor:main-dependent-rate}
Suppose
\begin{equation}
n\ge C\chi_X^2L
\max_{p\in\mathcal P_L,\,h\in\mathcal R_p}k_{p,h}.
\label{eq:rsc-sample-size}
\end{equation}
Then, after increasing the universal constant if necessary, with probability at least \(1-\delta\), simultaneously for all \((p,h,H)\),
\begin{align}
Q_p(\widehat B_{p,h,H}-B^\circ_{p,H})
\le{}& (2+64\chi_X)\mathcal A_{p,h,H} \nonumber\\
&+C\frac{\mathsf M_Xh}{\mathsf m_X^2}
\left\{
\frac{\mathsf S_{H,L}\Lambda_{H,L}q_p}{n}
+\frac{\Lambda_{H,L}^2q_p^2}{n^2}
\right\}. \label{eq:uniform-dependent-rate}
\end{align}
If additionally
\begin{equation}
n\ge C\max_{p,H}
\frac{\Lambda_{H,L}}{\mathsf S_{H,L}}q_p,
\label{eq:score-linear-regime}
\end{equation}
then the last display simplifies to
\begin{equation}
Q_p(\widehat B_{p,h,H}-B^\circ_{p,H})
\le (2+64\chi_X)\mathcal A_{p,h,H}
+C\frac{\mathsf M_Xh\mathsf S_{H,L}\Lambda_{H,L}}
{\mathsf m_X^2}\frac{q_p}{n}.
\label{eq:uniform-dependent-rate-linear}
\end{equation}
\end{corollary}

\begin{proof}
Condition \eqref{eq:rsc-sample-size} makes the right side of \eqref{eq:uniform-rsc-deviation} at most \(\mathsf m_X\|D\|_{\mathrm{F}}^2/2\).  Hence the restricted-curvature event holds with \(\kappa_p=\mathsf m_X/2\).  The same condition implies the fixed-oracle event \(\widehat Q_p(E_{p,h,H})\le2Q_p(E_{p,h,H})\). Theorem~\ref{thm:oracle}, with \(M_p=\mathsf M_X\), gives
\begin{equation}
Q_p(\widehat B_{p,h,H}-B^\circ_{p,H})
\le(2+64\chi_X)\mathcal A_{p,h,H}
+\frac{256\mathsf M_Xh}{\mathsf m_X^2}
\|G_{p,H}\|_{\mathrm{op}}^2.
\end{equation}
Substitute \eqref{eq:uniform-score} and use \((a+b)^2\le2a^2+2b^2\) to obtain \eqref{eq:uniform-dependent-rate}.  Condition \eqref{eq:score-linear-regime} absorbs the quadratic score term into the linear term.
\end{proof}

\paragraph{Interpretation.} The sample-size condition can be written as the one-sided implication
\begin{equation}
n_{\mathrm{curv}}^{\mathrm{suff}}(L)
\ge C\max_{p,h}k_{p,h}
\quad\Longrightarrow\quad
\text{the stated restricted curvature event}.
\end{equation}
This is only a sufficient curvature scale. It is not an information-equivalent sample size and no reverse implication is claimed.  The score bound separately contains the factor \(q_p=N_p+p+1+\ell_\delta\). The overlap cost is not asserted to be minimax sharp.  It arises from the worst-case Fourier concentration of an arbitrary low-rank reshaped lag filter. A sharper theorem would replace the envelope \(L\) by a localized Fourier-width functional of the rank class.

\subsection{Proof of the main finite-sample bound}
\label{app:main-estimation-proof}

\begin{proof}[Proof of Theorem~\ref{thm:patch-estimation}]
For each candidate $p$, apply Corollary~\ref{cor:uniform-dependent-rate} with rank $h=\min\{m,L/p,p\}$ and horizon $H$. This rank gives exactly the same coefficient class as $\mathcal{M}_m(p)$, including when the original rank constraint is inactive. Thus its approximation term is $A_m(p)$, and its population quadratic form is the squared prediction norm used here. There is one candidate rank per patch length, so its confidence term is $\log(12|\mathcal{P}_L|/\delta)$.

In this proof, the dimension formula can be written as $d_m(p)=h(L/p+p-h)$. Since $1\le h\le\min\{L/p,p\}$, both $h^2\le d_m(p)$ and $h\le d_m(p)$. Consequently,
\begin{align}
h(L/p+p+1)
 &=d_m(p)+h^2+h\le3d_m(p), \nonumber\\
 \min(2h,L/p,p)(L/p+p+1)
 &\le6d_m(p).
\end{align}
Also $L/p+p+1\le h(L/p+p+1)\le3d_m(p)$. These inequalities, together with $L,m\ge1$, show that \eqref{eq:patch-estimation-sample-size} implies both \eqref{eq:rsc-sample-size} and \eqref{eq:score-linear-regime} when $C_0$ is sufficiently large. The spectral factors entering these conditions depend only on the model, $H$, and $L$: reshaping preserves the population forecast residual and the eigenvalues of the context covariance.

The estimation term in \eqref{eq:uniform-dependent-rate-linear} is a fixed spectral prefactor times
\begin{equation}
 \frac{h}{n}
 \left\{\frac Lp+p+1+\log\frac{12|\mathcal{P}_L|}{\delta}\right\}.
\end{equation}
Since $h\le m$, this expression is bounded by
\begin{equation}
 \frac{3}{n}
 \left\{d_m(p)+m\log\frac{12|\mathcal{P}_L|}{\delta}\right\}.
\end{equation}
Absorb the factor three and the spectral prefactor into $C_2$, and take $C_1$ to be the approximation coefficient in that corollary. This proves \eqref{eq:patch-estimation-bound} simultaneously over all candidates with probability at least $1-\delta$. The constants are independent of $p$, $n$, $m$, $\mathcal{P}_L$, and $\delta$. The underlying concentration result accounts for overlap and serial dependence among training windows.

For an independent stationary test window, the projection identity \eqref{eq:risk}, conditional on the fitted coefficient, adds the common residual risk $R_0$ to this squared prediction-norm error. This proves the stated interpretation as a bound on prediction risk.
\end{proof}

\subsection{Squared and root risk under exact coefficient rank}
\label{app:discussion-integration}

We combine the finite-window bridge with the uniform estimation bound. Here $B^\circ_{p,H}=B^\circ_{H,L,p}$ and $\mathsf M_X$ is the observation spectral upper bound.

\begin{corollary}[Squared and root risk under exact coefficient rank]
\label{cor:exact-rank-root-rate}
Assume the finite-window bridge conditions of Theorem~\ref{thm:bridge}, the conditions of Corollary~\ref{cor:uniform-dependent-rate}, and the linear-score regime \eqref{eq:score-linear-regime}.  Fix a candidate \((p,h,H)\) for which the reshaped infinite-past kernel satisfies \(\operatorname{rank}(B_{H,p})\le h\).  Then, on the simultaneous event of Corollary~\ref{cor:uniform-dependent-rate},
\begin{align}
Q_p(\widehat B_{p,h,H}-B^\circ_{p,H})
\le{}&(2+64\chi_X)D_{H,L} \nonumber\\
&+C\frac{\mathsf M_Xh\mathsf S_{H,L}\Lambda_{H,L}}
{\mathsf m_X^2}\frac{q_p}{n}. \label{eq:exact-rank-square-rate}
\end{align}
Consequently,
\begin{align}
\sqrt{Q_p(\widehat B_{p,h,H}-B^\circ_{p,H})}
\le{}&\sqrt{2+64\chi_X}\,\sqrt{D_{H,L}} \nonumber\\
&+C\left\{
\frac{\mathsf M_Xh\mathsf S_{H,L}\Lambda_{H,L}}
{\mathsf m_X^2}\frac{q_p}{n}
\right\}^{1/2}. \label{eq:exact-rank-root-rate}
\end{align}
In particular, the finite-window contribution to root risk is \(O(\bar\rho_F^L)\), while the displayed estimation contribution is on the square root of the prediction-risk scale.
\end{corollary}

\begin{proof}
Because \(B_{H,p}\) itself is an admissible rank-\(h\) matrix, \(\mathcal E_h^\Gamma(B_{H,p})=0\).  The root-risk bridge \eqref{eq:bridge2} therefore gives
\begin{equation}
\sqrt{\mathcal A_{p,h,H}}
=\sqrt{\mathcal E_h^\Gamma(B^\circ_{H,L,p})}
\le\sqrt{D_{H,L}},
\end{equation}
and hence \(\mathcal A_{p,h,H}\le D_{H,L}\).  Substitution into \eqref{eq:uniform-dependent-rate-linear} proves \eqref{eq:exact-rank-square-rate}.  Apply \(\sqrt{a+b}\le\sqrt a+\sqrt b\) to obtain \eqref{eq:exact-rank-root-rate}.  Finally, \(D_{H,L}=\mathsf M_XK_H^2\bar\rho_F^{2L}/(1-\bar\rho_F^2)\) by Theorem~\ref{thm:bridge}.
\end{proof}

\subsection{Scope and possible refinements}

The logical chain is
\begin{align}
\text{stable Gaussian SSM}
&\Longrightarrow
(\mathsf m_X,\mathsf M_X,\mathsf M_{u,H,L}), \nonumber\\
(\mathsf m_X,\mathsf M_X,\mathsf M_{u,H,L})
&\Longrightarrow
\begin{cases}
\text{low-rank curvature},\\
\text{score concentration},\\
\text{fixed-oracle concentration},
\end{cases} \nonumber\\
\text{three events}
&\Longrightarrow\text{deterministic oracle inequality}
\Longrightarrow\text{uniform oracle rate}.
\end{align}
The concentration tools in Appendix~\ref{app:dependent-rate} establish the stated uniform prediction bound for global rank-constrained least-squares minimizers. The analysis does not provide a localized improvement of the worst-case \(L\) envelope, data-driven estimates of the spectral constants, an optimization guarantee for finding the global minimizer, or an oracle theorem for selecting \((p,h)\) from the same trajectory.

\paragraph{A possible localized refinement.} A sharper patch-dependent bound would require upper and lower bounds for
\begin{equation}
\gamma_2\!\left(
\{D:\operatorname{rank}(D)\le r,\ \|D\|_{\mathrm{F}}=1\},
(D,D')\mapsto\|a_{p,D-D'}\|_\infty
\right)
\end{equation}
on a localized secant or tangent class around the population optimizer. Establishing that this width is smaller than the global bound \(C\sqrt{Lr(N_p+p+1)}\) under verifiable spectral conditions on the coefficient matrix remains outside the present analysis. Neither Theorem~\ref{thm:uniform-dependent-events} nor Corollary~\ref{cor:uniform-dependent-rate} depends on such an improvement.

\section{PATCH SELECTION AND COMPUTATIONAL TRADEOFFS}
\label{app:patch-bounds}

The results below combine the approximation and estimation terms from the preceding sections. We first state a conditional selection criterion that allows a declared dependence adjustment and calculate the dimension of the admissible coefficient class, including the extreme cases. We then prove the selection statements from Section~\ref{sec:selection}, give a margin condition for the conditional criterion, and account for computational cost.

\subsection{When balanced patches are a useful reference}
\label{app:balanced-reference}

To state a heuristic that allows a user-specified dependence adjustment, we now write $h$ for a generic rank budget and $n_{\rm eff}(p)$ for a declared effective-sample proxy. For fixed rank bound $h$, the largest-dimensional part of the coefficient class has dimension
\begin{equation}
 d_p=\widetilde q_p(N+p-\widetilde q_p),\qquad
 \widetilde q_p=\min(h,N,p).
 \label{eq:dimension2}
\end{equation}
In the unsaturated case this is $h(L/p+p-h)$, minimized over continuous $p$ at $\sqrt L$. At either extreme, $p=1$ or $p=L$, every coefficient vector has a rank-one reshape, but the class still has $L$ free coefficients. Low rank alone therefore does not imply low complexity.

A simple fixed-regression comparison is
\begin{align}
\Psi(p)&=A_{H,L,h}(p)+c_{H,L}\frac{d_p}{n_{\rm eff}(p)}, \nonumber\\
A_{H,L,h}(p)&=\mathcal{E}_h^\Gamma(B^\circ_{H,L,p}), \label{eq:proxy3}
\end{align}
where $c_{H,L}>0$ is a penalty scale and $n_{\rm eff}(p)>0$ is a declared dependence proxy. The estimation theorem does not supply a universal choice of these quantities. When approximation error and the proxy are constant across patches, divisors near $\sqrt L$ minimize this criterion. Appendix~\ref{app:patch-insight} gives a margin condition allowing small approximation differences. When representation error changes substantially with $p$, it can outweigh the dimension advantage.

Nor is $d_p$ the parameter count of an attention network. The simulation control has separate query, positional-key, and value parameters, and two parameterizations of the same coefficient class can have different initialization, regularization, and optimization behavior. This distinction is not resolved by the fixed-regression bound and must be assessed empirically for the implemented architecture.

\subsection{Dimension and extreme patch lengths}

The manifold of real $N\times p$ matrices of exact rank $q$ has dimension $q(N+p-q)$: write $B=UV^\top$ and subtract the $q^2$-dimensional change-of-basis indeterminacy. The top attainable rank is $\widetilde q_p=\min\{h,N,p\}$, proving \eqref{eq:dimension2}. In the interior,
\begin{equation}
d_p=h(L/p+p-h).
\end{equation}
Thus minimizing dimension is equivalent to minimizing $L/p+p$, whose continuous minimizer is $\sqrt L$ by differentiation or the arithmetic--geometric mean inequality. For saturated candidates with $\min(N,p)\le h$, the dimension is $L$. For interior candidates it is $h(N+p-h)<L$. Thus a divisor minimizing $N+p$ is always dimension-minimizing. If every candidate is saturated, all dimensions equal $L$, so no unique preference follows.

\begin{lemma}[Rank one at extreme patch lengths is not compression]
\label{lem:patch-endpoints}
For a scalar series with context length \(L\) and any \(h\ge1\), both extreme reshapes \(p=1\) and \(p=L\) have matrix rank at most one.  Their rank-one approximation error is therefore zero, but the corresponding rank class has dimension \(L\) at both extremes.  Under dense token mixing, the analytic cost proxy is \(L^2\) at \(p=1\) and \(1\) at \(p=L\).
\end{lemma}

\begin{proof}
At \(p=1\), the coefficient matrix is \(L\times1\). At \(p=L\), it is \(1\times L\).  Each has rank at most one.  With attainable rank \(q=1\), the rank-class dimension is \(q(N+p-q)=N+p-1=L\) in either case.  The dense nonoverlapping token-mixing proxy is \(N^2=(L/p)^2\), which gives the two stated values.
\end{proof}

\subsection{A sufficient condition from the forecast kernel}
\label{app:patch-kernel-condition}

The filtering analysis provides a structural sufficient condition for the theorem. By Theorem~\ref{thm:factor}, $m\ge r$ makes the truncated infinite-past kernel $B_{H,p}$ representable at every patch length. This does not imply that the optimal finite-window coefficient $B_p^\circ$ has rank at most $m$. The relevant question is instead whether the discrepancy between these two coefficients is small enough to preserve the estimation advantage of balance.

\begin{corollary}[A sufficient condition from the forecast kernel]
\label{cor:patch-finite-window}
Maintain the conditions of Theorems~\ref{thm:patch-estimation}, \ref{thm:factor}, and~\ref{thm:bridge}. Suppose $r\le m<\min\{L/q,q\}$ for every $q\in\mathcal{P}_L$, and let $p\in\mathcal{P}_L$ minimize $d_m(p)$. If
\begin{equation}
 C_1\|B_p^\circ-B_{H,p}\|_\Gamma^2
 <\frac{C_2}{n}\{d_m(q)-d_m(p)\}
 \label{eq:patch-finite-window-condition}
\end{equation}
for every $q\in\mathcal{P}_L$ with $d_m(q)>d_m(p)$, then every minimizer of \eqref{eq:proxy} minimizes $p+L/p$ over $\mathcal{P}_L$.
\end{corollary}

The corollary follows because the representable kernel is an admissible approximation to the finite-window target:
\begin{equation}
 0\le A_m(p)\le\|B_p^\circ-B_{H,p}\|_\Gamma^2.
\end{equation}
Its condition therefore requires the cost of using this approximation to be smaller than the estimation saving, even relative to a higher-dimensional candidate with no approximation error. Theorem~\ref{thm:bridge} bounds the discrepancy geometrically in $L$. The displayed condition specifies how small it must be for the comparison at hand. Exact low rank of the finite-window target is unnecessary.

Theorem~\ref{thm:patch-selection} does not require $m\ge r$. With fewer heads, however, the preceding sufficient argument is unavailable, and patch choice must account for the approximation errors described by the filtering spectrum and \eqref{eq:bridge2}. A single dominant forecast component does not by itself remove the filtering patterns needed to recover that component from past observations. Nor does having fewer heads imply that patch length is immaterial: its effect depends on how well the reshaped target can be approximated.

\subsection{Proofs of the main patch-selection results}

\begin{proof}[Proof of Theorem~\ref{thm:patch-selection}]
On the stated candidate set, minimizing $d_m(p)$ is equivalent to minimizing $p+L/p$. To prove the theorem, the difference between the criterion at $q$ and at a dimension-minimizing $p$ is
\begin{equation}
 C_1\{A_m(q)-A_m(p)\}
 +\frac{C_2}{n}\{d_m(q)-d_m(p)\}.
\end{equation}
Under \eqref{eq:patch-selection-margin}, every candidate of higher dimension has a strictly larger criterion than $p$. Moreover, $p+L/p\ge2\sqrt L$, with equality only at $p=\sqrt L$. Hence $\sqrt L$, when admissible, is the unique dimension minimizer. Rearranging the inequality between the criteria at $p$ and $q$ gives \eqref{eq:patch-sample-transition}, with equality corresponding to a tie.
\end{proof}

\begin{proof}[Proof of Corollary~\ref{cor:patch-finite-window}]
By Theorem~\ref{thm:factor}, $B_{H,p}\in\mathcal{M}_m(p)$ when $m\ge r$. Feasibility and nonnegativity of approximation error therefore give $A_m(p)-A_m(q)\le\|B_p^\circ-B_{H,p}\|_\Gamma^2$. The stated condition therefore implies \eqref{eq:patch-selection-margin}.
\end{proof}

\subsection{A margin condition for balanced patches}
\label{app:patch-insight}

Fix $H,L,h$ and abbreviate $A(p)=A_{H,L,h}(p)$. Let $p_{\rm bal}$ minimize $d_p$ over the candidate set. Assume $c_{H,L}>0$ and a patch-independent proxy $n_{\rm eff}>0$. If every candidate has the same dimension, the margin below is unnecessary. Otherwise the minimum positive dimension gap in the statement is well-defined.

\begin{lemma}[Patch-selection margin]
\label{lem:patch-margin}
Let $p_{\rm bal}$ minimize $d_p$ over a finite candidate set $\mathcal{P}_L$ with at least two distinct dimensions. If
\begin{equation}
\sup_p A(p)-\inf_p A(p)
<c_{H,L}\frac{\min_{d_p>d_{p_{\rm bal}}}(d_p-d_{p_{\rm bal}})}{n_{\mathrm{eff}}},
\end{equation}
then every minimizer of $\Psi$ over $\mathcal{P}_L$ belongs to the dimension-minimizing set.
\end{lemma}

\begin{proof}
Let \(p_0\) be any dimension minimizer. For every \(p\) with \(d_p>d_{p_0}\),
\begin{equation}
\Psi(p)-\Psi(p_0)
\ge
-\left\{\sup_{q\in\mathcal{P}_L}A(q)-\inf_{q\in\mathcal{P}_L}A(q)\right\}
+\frac{c_{H,L}}{n_{\mathrm{eff}}}(d_p-d_{p_0})>0.
\end{equation}
Thus no candidate outside the dimension-minimizing set can minimize \(\Psi\).
\end{proof}

\subsection{Computational-cost tradeoffs}
\label{app:cost-details}
Let $a>0$ collect the statistical penalty scale and let $c_{\rm dense}\ge0$ price dense token mixing. Suppressing approximation error gives
\begin{equation}
 \Phi(p)=a(L/p+p)+c_{\rm dense}L^2/p^2.
 \label{eq:compute}
\end{equation}
Its positive continuous minimizer solves
\begin{equation}
 p^3-Lp-2(c_{\rm dense}/a)L^2=0.
 \label{eq:cubic}
\end{equation}
For fixed $m,d_k,d_v$, the dot-product attention contraction instead costs $O(mNd_kd_v)$. For $c_{\rm lin}\ge0$, a corresponding linear-mixing proxy is
\begin{equation}
 \Phi_{\rm lin}(p)=a(L/p+p)+c_{\rm lin}L/p.
 \label{eq:linear-compute}
\end{equation}
Its continuous minimizer is $\sqrt{L(1+c_{\rm lin}/a)}$. These formulas show why the computational part of patch selection must match the architecture. Evaluate the admissible patch lengths directly, since not every integer divides $L$. The formulas omit projection and encoder costs. They do not predict total running time or neural-network test error. Softmax attention with a single query also has mixing cost linear in $N$. The rank-based penalty is a fixed-regression heuristic, not a count of learned attention capacity. For learned attention, use the validation procedure in Appendix~\ref{sec:discussion-implications} and report the full prediction-error versus cost comparison, not only the selected patch.

\begin{proof}[Derivation of the cost minimizers]
For the compute proxy \eqref{eq:compute}, differentiation yields
\begin{equation}
\Phi'(p)=a\left(1-\frac{L}{p^2}\right)
-2c_{\rm dense}\frac{L^2}{p^3}.
\end{equation}
Multiplying by $p^3/a$ gives \eqref{eq:cubic}. When $c_{\rm dense}>0$, the cubic is negative at $p=\sqrt L$ and has a unique positive zero above it. If $p^3$ dominates $Lp$, balancing with $2(c_{\rm dense}/a)L^2$ gives the $L^{2/3}$ regime.

For the linear-mixing proxy \eqref{eq:linear-compute}, the derivative vanishes at $p^2=L(1+c_{\rm lin}/a)$ and the second derivative is $2(a+c_{\rm lin})L/p^3>0$. Restricting to $1\le p\le L$ clips the continuous optimum. A divisor constraint requires evaluating the discrete candidates. This calculation includes neither a learned query/key complexity term nor an approximation curve.
\end{proof}

\subsection{Interpretation and validation}

The theory does not say that a longer patch is always a stronger token, or that \(p=\sqrt L\) is universally optimal. A patch token is the vector \((X_{t-jp},\ldots,X_{t-jp-p+1})\). Its length \(p\) determines how much local temporal structure is packaged together, while \(N=L/p\) determines how many tokens must interact. Table~\ref{tab:patch-selection-forces} summarizes the competing forces. Here \(n_{\mathrm{eff}}(p)\) denotes a declared dependence-adjustment proxy used only in the conditional selector. It is not asserted to equal the one-sided sufficient curvature scale \(n/(\chi_X^2L)\) proved in Appendix~\ref{app:dependent-rate}.

\begin{table}[htbp]
\centering
\caption{Approximation, estimation, and computational considerations in patch selection.}
\label{tab:patch-selection-forces}
\begin{tabular}{@{}p{0.18\textwidth}p{0.38\textwidth}p{0.38\textwidth}@{}}
\toprule
Force & Quantity represented in the theory & Typical pressure on \(p\) \\
\midrule
Predictive approximation & \(A_{H,L,h}(p)\) or its multi-horizon aggregate & Depends on the coefficient reshape under a rank constraint. Patching alone is lossless. No monotone direction is proved. \\
Statistical complexity & \(d_p/n_{\mathrm{eff}}(p)\), with \(d_p=h(L/p+p-h)\) when \(h\le\min(N,p)\) & Toward a balanced factorization near \(\sqrt L\) only when approximation and the dependence proxy vary little. \\
Serial dependence & Patch-dependent score concentration and restricted curvature & Smaller \(n_{\mathrm{eff}}\) magnifies complexity control. The direction is not universally larger because \(n_{\mathrm{eff}}(p)\) can change with tokenization. \\
Token mixing & \((L/p)^2\) for dense all-token mixing, and \(L/p\) for fixed-width linear attention & Larger patches reduce mixing. The \(L^{2/3}\) scale belongs only to the dense quadratic proxy. \\
Multiple horizons & Weighted-average or worst-case approximation over \(H\in\mathcal H\) & One shared patch only if horizon-specific curves share a near-minimizer. Otherwise multiresolution or horizon-specific tokens. \\
\bottomrule
\end{tabular}
\end{table}

At \(p=1\), each scalar observation is a token and \(N=L\). The matrix \(B_{H,1}\) is necessarily rank one because it has one column, so a zero rank-one approximation error carries no compression content. These facts identify two costs of pointwise tokenization: it leaves all $L$ coefficients free and maximizes dense token mixing. The separate weak-token hypothesis concerns the local structure that learned attention scores must recover. It is an architectural consideration, not a theorem that scalar tokens are intrinsically uninformative.

For implementation, use the candidate divisors of \(L\), estimate forecast loss with chronological validation and guard gaps, and prespecify how computational cost enters the comparison. Report the entire patch-risk curve as well as the selected value. The curve is scientifically more informative than a single optimizer because it reveals whether the choice is driven by approximation, statistical uncertainty, or compute. For multiple horizons, prespecify the horizon weights or worst-case criterion rather than choosing them after observing test performance.

\section{CONTEXT ADAPTATION ACROSS SYSTEMS}
\label{app:context-mixture-details}

For one known Gaussian system, the optimal finite-window forecast is already linear, so input-dependent attention weights cannot lower the population Bayes risk. One setting in which adaptation can be useful is when the observed context reveals which forecasting rule should be used. A mixture of systems provides a simple example: the history updates the posterior probability of each system, and those posterior probabilities mix the corresponding linear forecasts.

Let $Y=X_{t+H}$ be the forecast target. For any square-integrable forecast $g(Z)$, conditional expectation gives
\begin{equation}
\mathbb E(Y-g(Z))^2=\mathbb E(Y-f^*(Z))^2+\mathbb E(g(Z)-f^*(Z))^2.
\end{equation}
Here $f^*(Z)=\mathbb E(Y\mid Z)$ in the model under consideration. Thus adaptation has a strict population advantage over the best fixed linear predictor exactly when that linear predictor differs from $f^*$ with positive probability. Merely varying the weights does not establish this advantage.

\subsection{Mixture model and posterior weights}

Let $M$ be the number of generating systems and draw a latent index $U\in\{1,\ldots,M\}$ once per independent episode, with prior probabilities $\omega_u=\Pr(U=u)>0$ summing to one. Conditional on $U=u$, generate a stationary episode from a specified stable mean-zero Gaussian system. There is no switching within an episode. Let $\ell_t=(X_t,\ldots,X_{t-L+1})^\top$ have conditional covariance $G_u\succ0$, and let $B_u^\circ$ be the optimal finite-window coefficient matrix in system $u$. These coefficients need not have the same rank as the truncated infinite-past kernels. Writing $\mathcal N(\ell;0,G)$ for the centered Gaussian density with covariance $G$, the posterior is
\begin{equation}
 \pi_u(Z)=\frac{\omega_u\,\mathcal N(\ell_t;0,G_u)}
 {\sum_s\omega_s\,\mathcal N(\ell_t;0,G_s)}
\,.
\end{equation}
This probability uses observed inputs only, not the future target. Identical past covariances give constant posteriors, so context adaptation is not guaranteed for arbitrary choices of the systems.

\subsection{Oracle forecast and rank truncation}

\begin{theorem}[Finite-mixture oracle and spectral truncation]
\label{thm:context-mixture}
Under the preceding assumptions, the Bayes forecast is $f^*(Z)=\sum_u\pi_u(Z)\langle B_u^\circ,Z\rangle_{\mathrm{F}}$. Choose deterministic best rank-$h$ Frobenius approximations $B_{u,h}^\circ$ and define $f_h^{\rm mix}(Z)=\sum_u\pi_u(Z)\langle B_{u,h}^\circ,Z\rangle_{\mathrm{F}}$. Writing $\Gamma_u=\mathbb E[\operatorname{vec}(Z)\operatorname{vec}(Z)^\top\mid U=u]$,
\begin{align}
\mathbb E(Y-f_h^{\rm mix}(Z))^2-\mathbb E(Y-f^*(Z))^2
 &=\mathbb E(f_h^{\rm mix}(Z)-f^*(Z))^2 \nonumber\\
&\le\sum_u\omega_u\lambda_{\max}(\Gamma_u)
              \sum_{k>h}\sigma_k(B_u^\circ)^2. \label{eq:mixture-tail-bound}
\end{align}
\end{theorem}
\begin{proof}
Conditional Gaussian regression gives $\mathbb E(Y\mid Z,U=u)=\langle B_u^\circ,Z\rangle_{\mathrm{F}}$. The tower property gives $f^*$. Orthogonality of $Y-f^*(Z)$ to any square-integrable function of $Z$ gives the excess-risk identity. Set $\Delta_u=B_{u,h}^\circ-B_u^\circ$. Conditional Jensen yields
\begin{equation}
 \left(\sum_u\pi_u(Z)\langle \Delta_u,Z\rangle_{\mathrm{F}}\right)^2
 \le\sum_u\pi_u(Z)\langle \Delta_u,Z\rangle_{\mathrm{F}}^2.
\end{equation}
For any integrable $g(Z)$, $\mathbb E[\pi_u(Z)g(Z)]=\omega_u\mathbb E[g(Z)\mid U=u]$. Apply this identity, then bound the conditional quadratic form by $\lambda_{\max}(\Gamma_u)\|\Delta_u\|_{\mathrm{F}}^2$ and use truncated SVD.
\end{proof}

\subsection{Implications for adaptive attention}

This is an oracle mixture approximation result, not a claim that one dot-product attention layer exactly implements Gaussian posterior inference. The $\pi_u(Z)$ here are posterior probabilities over systems. They are not the signed patch coefficients $\alpha_{H,a,j}(Z)$. The effective mixture coefficient $\sum_u\pi_u(Z)B_{u,h}^\circ$ has rank at most $Mh$, which can exceed $h$. Shared right-factor subspaces can reduce this upper bound. For a learned $\widehat f$, its squared error to $f^*$ is at most twice the right side of \eqref{eq:mixture-tail-bound} plus $2\mathbb E|\widehat f(Z)-f_h^{\rm mix}(Z)|^2$. Controlling the latter requires an approximation and learning analysis of the actual architecture. The single-system, one-trajectory rate in Appendix~\ref{app:dependent-rate} is not a theorem for the non-Gaussian marginal mixture or for neural training.

\section{DISCUSSION AND IMPLICATIONS}
\label{sec:discussion-implications}

This section collects practical implications of the results: when context-dependent attention weights can help, and how patch lengths and head counts should be compared in practice.

\subsection{When context dependence can help}
\label{sec:adaptation}

For a known Gaussian system, the optimal finite-window forecast is linear. Context dependence can become useful when each episode is generated by an unknown system and the observed window contains information about its identity. For a fixed patch length, let $U\in\{1,\ldots,M\}$ index the generating system, let $B_u^\circ$ be its optimal finite-window coefficient matrix, and write $\pi_u(Z)=\Pr(U=u\mid Z)$. If each system is centered and Gaussian, the Bayes forecast under squared loss is
\begin{equation}
 f^*(Z)=\sum_{u=1}^M\pi_u(Z)\langle B_u^\circ,Z\rangle_{\mathrm{F}}.
 \label{eq:mixture-bayes}
\end{equation}
The posterior weights generally vary with $Z$, so this forecast can be nonlinear even though the forecast within each system is linear. Theorem~\ref{thm:context-mixture} in Appendix~\ref{app:context-mixture-details} bounds the approximation error when these posterior weights are available. Learning the weights and representing the resulting forecast require a separate analysis of the attention architecture. Fixed-system experiments alone cannot establish this potential benefit of adaptation across systems.

\subsection{Computational cost and validation}

Computational cost introduces a separate consideration. At fixed widths and head count, dense all-token mixing costs $O(N^2)$, whereas associative unnormalized attention and single-query attention have mixing costs of order $N$, where $N=L/p$. Larger patches reduce these mixing costs, but the total cost also includes patch projections and any context encoder. A computational penalty that varies with $p$ changes the objective in \eqref{eq:proxy} and may change its minimizer. Appendix~\ref{app:cost-details} gives illustrative mixing-cost calculations. Runtime and memory should also be measured for the implemented model.

In practice, compare patch lengths and head counts using common training and validation splits, comparable tuning budgets, and a fixed linear or positional-pooling control. For a single trajectory, use chronological splits with a gap that prevents training and validation windows and targets from overlapping. For independent series or episodes, hold out entire series or episodes. Select patch length, head count, checkpoints, and other tuning parameters using validation data. Prespecify how prediction error is traded against computational cost and how results are aggregated across horizons, and reserve the test set for final evaluation.

Pointwise tokenization ($p=1$) can produce weak tokens for learned attention: queries and keys formed from individual observations do not directly encode local temporal patterns. Empirical ablations demonstrate benefits from replacing such tokens with patches \citep{nie2023patchtst}. This is compatible with $A_m(1)=0$, which concerns the representational capacity of the fixed linear class, rather than the learning of informative attention scores. Weak-token effects thus provide an additional architectural motivation for patching that is not captured by the present fixed-weight estimation bound.

\section{SIMULATION DIAGNOSTICS AND REPRODUCIBILITY}
\label{app:experiments}

This section separates calculations that depend only on the known state-space model from errors introduced by finite samples and model fitting. All population quantities use the same lag ordering as the main paper. Random seeds, candidate grids, optimizer settings, stopping rules, the number of windows per series, and the computing environment are listed in Appendix~\ref{app:add-protocol}.

\subsection{Design of the main simulations}
\label{app:sim-design}

The simulations in Section~\ref{sec:experiments} use the three-state model \eqref{eq:ssm} with
\begin{equation}
 A=\operatorname{diag}(0.95,0.88,0.80),\qquad C=(1,0.7,0.5).
 \label{eq:sim-AC}
\end{equation}
The latent components have different persistence levels, and each observation combines all three states with measurement noise. Process and measurement noises are independent Gaussian draws. Starting from process covariance $I_3$ and measurement variance $0.3$, we rescale both so that the stationary observation variance is one (Appendix~\ref{app:sim-population}), and we initialize each independent path from its stationary distribution.

We consider window lengths $L\in\{64,144\}$, horizons $H\in\{1,2,4,8,12\}$, and training sets of $n\in\{5,25,200\}$ independent series. Unlike the theoretical $n$, which counts consecutive windows from one trajectory, this $n$ counts independent series. Each series supplies overlapping input--target windows, and entire series are kept separate across the training, validation, and test sets. For each window length, the patch grid contains the balanced reference $p=\sqrt L$, nearby divisors, and the extremes $p=1$ and $p=L$. Appendix~\ref{app:add-protocol} lists the training protocol.

\subsection{Population construction}
\label{app:sim-population}

Begin with the transition and observation matrices in \eqref{eq:sim-AC}, process covariance $I_3$, and measurement variance $0.3$. If $P$ solves
\begin{equation}
 P=APA^\top+I_3,
\end{equation}
then the unscaled stationary observation variance is $v= CPC^\top+0.3$. Dividing both noise covariances by $v$ gives
\begin{equation}
 Q=I_3/v,\qquad R=0.3/v.
\end{equation}
Thus $\operatorname{Var}(X_t)=1$. This common rescaling of the process and measurement noise does not change the steady-state Kalman gain. The predictor gain $K$ and $F=A-KC$ are obtained from the stabilizing discrete Riccati solution under the predictor convention in Equation~\eqref{eq:observer}.

For an independent check that does not rely on the Kalman recursion, let $\gamma_X(k)=\operatorname{Cov}(X_{t+k},X_t)$. The exact finite-window coefficient $b^\circ_{H,L}$ solves
\begin{equation}
 \Gamma_L b^\circ_{H,L}=g_{H,L},\qquad
 [\Gamma_L]_{ij}=\gamma_X(i-j),\quad
 [g_{H,L}]_j=\gamma_X(H+j),
 \label{eq:simulation-population-normal-equations}
\end{equation}
for $0\le i,j<L$. Reshaping this vector gives $B^\circ_{H,L,p}$ for every candidate $p\mid L$. The population rank-$h$ approximation error is then
\begin{equation}
 A_{H,L,h}(p)=
 \min_{\operatorname{rank}(B)\le h}
 \|B-B^\circ_{H,L,p}\|_{\Gamma_{L,p}}^2.
\end{equation}
These calculations distinguish approximation error from the error of a trained network.

\subsection{Population filtering-floor diagnostic}
\label{app:simulation-floor}

For each $p$ and horizon $H$, form the truncated infinite-past matrix $B_{H,p}$ from Equation~\eqref{eq:patchmatrix}. For a one-head model, report both the absolute prediction-norm error $\mathcal{E}_1^\Gamma(B_{H,p})$ and the relative spectral diagnostic
\begin{equation}
 \mathcal R_{H,p}
 =\frac{\sigma_2(B_{H,p})}{\sigma_1(B_{H,p})},
 \label{eq:simulation-relative-floor}
\end{equation}
whenever $B_{H,p}\ne0$. The first quantity can shrink with $H$ as the predictable signal decays, whereas Theorem~\ref{thm:floor} allows the second to converge to a positive limit. Plotting both prevents a relative filtering floor from being mistaken for a nonvanishing absolute forecast error. The finite-window analogue replaces $B_{H,p}$ by $B^\circ_{H,L,p}$ computed from \eqref{eq:simulation-population-normal-equations}.

\subsection{Fitted-model protocol and robustness sweeps}
\label{app:simulation-robustness}

Each repetition begins with newly simulated stationary series. Entire series, rather than overlapping windows, are assigned to training, validation, or test sets. All patch candidates within a repetition use the same simulated series, splits, initialization schedule, tuning budget, and checkpoint rule. Any standardization beyond the population scaling above is computed from the training split and then held fixed. Validation data select the checkpoint and any optimization hyperparameters. Test data are used once for the reported comparison.

The horizon sweep holds the number of training series fixed and evaluates every $H\in\{1,2,4,8,12\}$. The training-size sweep holds the data-generating model, window length, horizon, and per-series window construction fixed while varying the number of independent training series. For each setting, report all admissible patch candidates, not only $p=1$, $p=\sqrt L$, and $p=L$. Excess MSE is evaluated against the analytic conditional mean obtained from \eqref{eq:simulation-population-normal-equations}. Total MSE is evaluated against the realized future observation. Pointwise bootstrap intervals resample complete data repetitions, not individual overlapping windows, and table standard errors are computed across those repetitions.

Because $n$ in these simulations counts independent series while $n$ in the one-trajectory bound counts consecutive windows, the experiments test the qualitative approximation--estimation tradeoff rather than calibrating the constants in Theorem~\ref{thm:patch-estimation}. The exact settings are listed in Appendix~\ref{app:add-protocol}.

\section{ADDITIONAL EXPERIMENTS}
\label{app:additional-experiments}

This section reports further simulations with the fixed attention model of Section~\ref{sec:experiments}. Appendix~\ref{app:add-protocol} lists the training protocol shared by all runs. Appendices~\ref{app:add-aligned}--\ref{app:add-floor} complement the main simulations. Appendices~\ref{app:add-heads}--\ref{app:add-scaling} vary the head count, the rank of the target, and the window length. Unless stated otherwise, \emph{population excess risk} is $\|\widehat B_{p,m,H}-B^\circ_{H,L,p}\|_{\Gamma_{L,p}}^2$, computed exactly from the true model for each trained coefficient matrix, and intervals are pointwise $95\%$ bootstrap intervals that resample complete data repetitions.

\subsection{Training protocol and computing environment}
\label{app:add-protocol}

All runs train the fixed model of Section~\ref{sec:experiments}: an input-independent query, free positional keys, shared linear values and output weights without bias, and key and value width $8$ per head. Training uses AdamW with batch size $64$ and weight decay $10^{-4}$, at most $4{,}000$ updates, validation every $100$ updates, and early stopping with patience $10$. The learning rate is chosen from $\{0.001,0.003,0.015\}$ and the checkpoint by validation MSE. Every patch length receives the same search and update budget, and test data are never used for selection.

For the simulations of Section~\ref{sec:experiments} and Appendices~\ref{app:add-aligned}--\ref{app:add-repr}, each data repetition draws stationary series of length $576$ with $32$ forecast windows per series, holds out $50$ validation and $100$ test series, and uses nested training sets of $n\in\{5,25,200\}$ series. There are $50$ data repetitions (seeds $10,20,\ldots,500$) and one optimization seed. The patch grids are $\{1,2,4,8,16,32,64\}$ for $L=64$ and $\{1,2,6,12,24,72,144\}$ for $L=144$. These runs used PyTorch~2.5.1 in single precision on a CPU (Apple silicon, six threads).

The studies in Appendices~\ref{app:add-heads}--\ref{app:add-scaling} use the same model, optimizer, and selection rules with $100$ training, $50$ validation, and $100$ test series of length $6{,}400$ per repetition, $32$ forecast origins per series shared across all settings, $20$ data repetitions (seeds $3010,3020,\ldots,3200$), and every divisor of $L$ as a candidate patch length.

\subsection{A second observation design}
\label{app:add-aligned}

Section~\ref{sec:experiments} uses the dense observation vector $C=(1,0.7,0.5)$, which mixes all three latent states. We repeat the complete experiment with the near-aligned vector $C=(1,0.05,0.02)$, keeping $A=\operatorname{diag}(0.95,0.88,0.80)$ and rescaling the noises so that $\operatorname{Var}(X_t)=1$. The observation is then dominated by the most persistent state, and the one-step Bayes MSE is $0.144$, against $0.170$ for the dense design.

\begin{figure}[!htbp]
\centering
\includegraphics[width=0.9\textwidth]{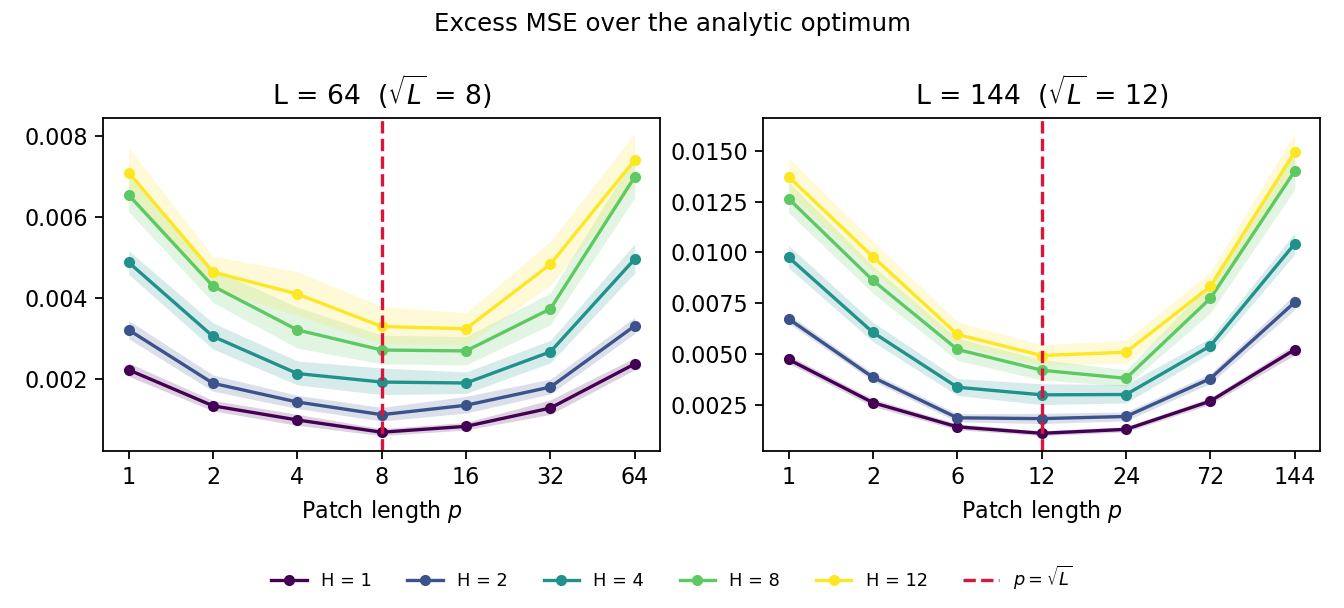}\\[0.5em]
\includegraphics[width=0.9\textwidth]{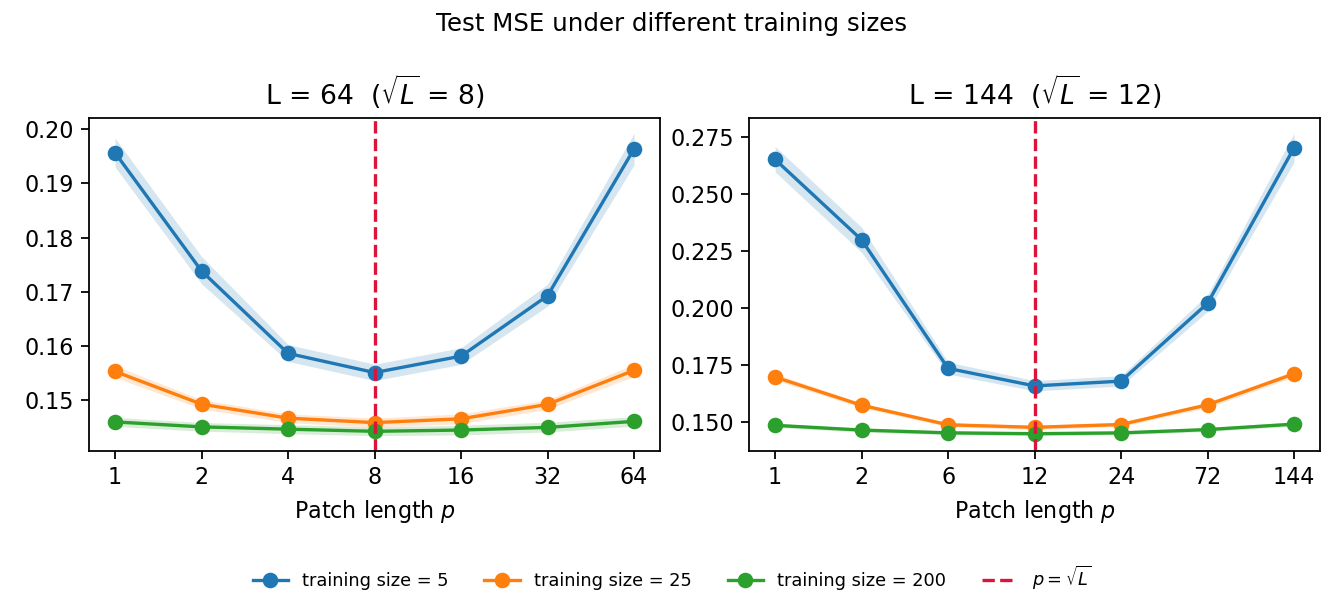}
\caption{Near-aligned observation design, fixed one-head attention, $50$ data repetitions. Top: excess MSE across forecast horizons with $n=200$ training series. Bottom: total test MSE at $H=1$ as the number of training series grows. Conventions match Figures~\ref{fig:simulation-horizons} and~\ref{fig:simulation-training}.}
\label{fig:add-aligned}
\end{figure}

Figure~\ref{fig:add-aligned} and Table~\ref{tab:add-aligned} reproduce the pattern of Section~\ref{sec:experiments}. With $n=200$, the patch length with the lowest population excess risk is $\sqrt L$ or the next larger divisor at every horizon and both window lengths, and both extremes have $2.1$ to $4.8$ times the excess risk at $\sqrt L$. The interior advantage is largest when data are scarce: at $L=144$ and $n=5$, the one-step test MSE is $165.6\times10^{-3}$ at $p=\sqrt L$ against $265.0\times10^{-3}$ and $270.1\times10^{-3}$ at the extremes, and the gap narrows as $n$ grows.

\begin{table}[!htbp]
\centering
\caption{Near-aligned observation design: one-step test MSE in units of $10^{-3}$ for the balanced patch and both extremes (means $\pm$ standard errors over $50$ data repetitions). The one-step Bayes MSE is $144$ in these units. Compare Table~\ref{tab:simulation-main}.}
\label{tab:add-aligned}
\begin{tabular}{@{}rrccc@{}}
\toprule
$L$ & $n$ & $p=1$ & $p=\sqrt L$ & $p=L$\\
\midrule
64 & 5 & $195.6 \pm 1.4$ & $155.1 \pm 0.8$ & $196.4 \pm 1.5$\\
64 & 25 & $155.3 \pm 0.5$ & $145.9 \pm 0.4$ & $155.5 \pm 0.6$\\
64 & 200 & $146.0 \pm 0.4$ & $144.3 \pm 0.4$ & $146.1 \pm 0.5$\\
\midrule
144 & 5 & $265.0 \pm 3.0$ & $165.6 \pm 1.2$ & $270.1 \pm 3.3$\\
144 & 25 & $169.5 \pm 0.7$ & $147.4 \pm 0.5$ & $170.9 \pm 0.7$\\
144 & 200 & $148.3 \pm 0.5$ & $144.6 \pm 0.4$ & $148.8 \pm 0.5$\\
\bottomrule
\end{tabular}
\end{table}

\subsection{Representation versus estimation in the main simulations}
\label{app:add-repr}

The population quantities of Appendix~\ref{app:experiments} separate the two error sources directly. Table~\ref{tab:add-repr} compares an upper bound on the one-head approximation error, obtained from the truncated singular value decomposition of $B^\circ_{H,L,p}$ and evaluated in the prediction norm, with the population excess risk of the trained model at $p=\sqrt L$. Even with $n=200$ training series the approximation bound is at most $7\%$ of the trained excess risk for the dense design, and it is negligible for the near-aligned design. At the extremes the approximation error is exactly zero. The differences between patch lengths in Figures~\ref{fig:simulation-horizons}, \ref{fig:simulation-training} and~\ref{fig:add-aligned} are therefore driven by estimation rather than representation, as the dimension term in \eqref{eq:proxy} predicts.

\begin{figure}[!htbp]
\centering
\includegraphics[width=0.9\textwidth]{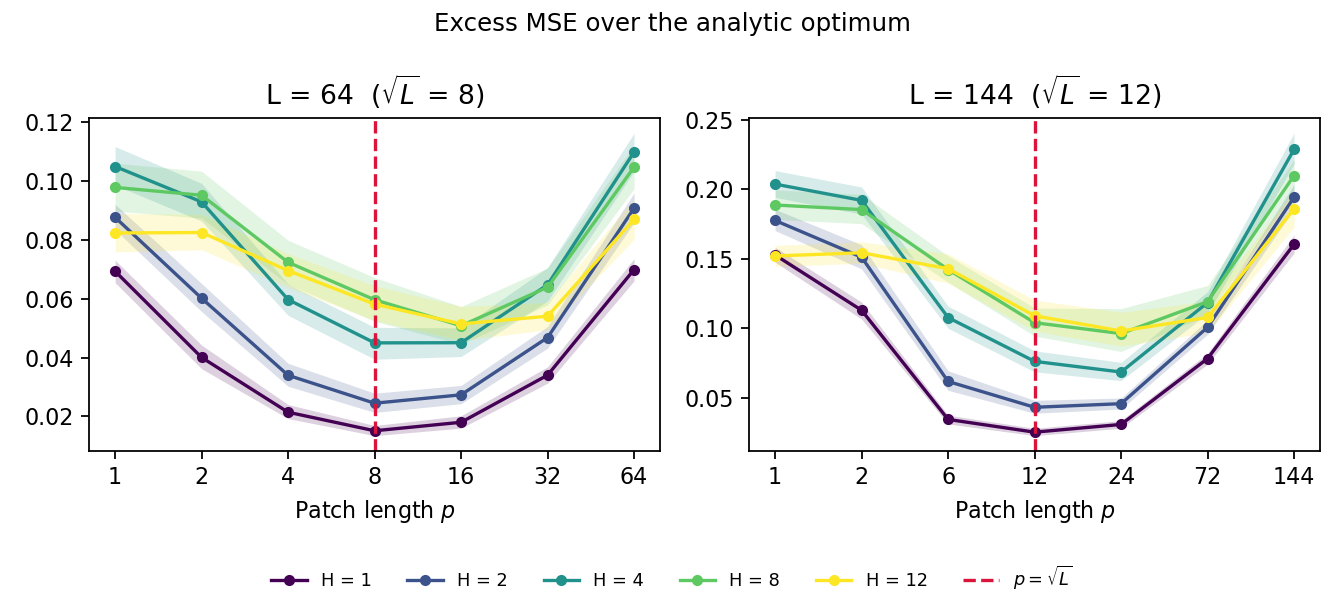}
\caption{Dense observation design with $n=5$ training series: excess MSE across forecast horizons for fixed one-head attention ($50$ repetitions). Conventions match Figure~\ref{fig:simulation-horizons}.}
\label{fig:add-dense-n5}
\end{figure}

Figure~\ref{fig:add-dense-n5} shows the scarce-data case for the dense design. With five training series, the extremes have $1.4$ to $6.3$ times the excess risk at $\sqrt L$. The lowest risk is at $p=\sqrt L$ for short horizons and at the next larger divisor for longer horizons, again consistent with a broad interior range rather than a unique optimum.

\begin{table}[!htbp]
\centering
\caption{One-head approximation error versus trained excess risk. ``Approx.\ bound'' is an upper bound on $\mathcal E_1^\Gamma(B^\circ_{H,L,p})$ at $p=\sqrt L$, and ``max over $p$'' its largest value over the patch grid. ``Trained'' is the mean population excess risk of the fitted model at $p=\sqrt L$ ($50$ repetitions).}
\label{tab:add-repr}
\begin{tabular}{@{}lrrcccccc@{}}
\toprule
Design & $L$ & $H$ & Signal & Approx.\ bound & Max over $p$ & Trained, $n=5$ & Trained, $n=200$\\
\midrule
Dense & 64 & 1 & 0.830 & $2.3\cdot10^{-5}$ & $4.6\cdot10^{-5}$ & $1.50\cdot10^{-2}$ & $9.2\cdot10^{-4}$\\
Dense & 64 & 12 & 0.206 & $2.6\cdot10^{-4}$ & $4.7\cdot10^{-4}$ & $5.80\cdot10^{-2}$ & $3.8\cdot10^{-3}$\\
Dense & 144 & 1 & 0.830 & $9.3\cdot10^{-6}$ & $3.5\cdot10^{-5}$ & $2.52\cdot10^{-2}$ & $1.3\cdot10^{-3}$\\
Dense & 144 & 12 & 0.206 & $1.1\cdot10^{-4}$ & $3.8\cdot10^{-4}$ & $1.08\cdot10^{-1}$ & $5.9\cdot10^{-3}$\\
\midrule
Near-aligned & 64 & 1 & 0.856 & $8.2\cdot10^{-10}$ & $1.6\cdot10^{-9}$ & $1.12\cdot10^{-2}$ & $6.9\cdot10^{-4}$\\
Near-aligned & 64 & 12 & 0.277 & $1.0\cdot10^{-8}$ & $2.0\cdot10^{-8}$ & $5.40\cdot10^{-2}$ & $3.3\cdot10^{-3}$\\
Near-aligned & 144 & 1 & 0.856 & $3.0\cdot10^{-10}$ & $1.3\cdot10^{-9}$ & $2.15\cdot10^{-2}$ & $1.1\cdot10^{-3}$\\
Near-aligned & 144 & 12 & 0.277 & $3.9\cdot10^{-9}$ & $1.6\cdot10^{-8}$ & $9.16\cdot10^{-2}$ & $4.9\cdot10^{-3}$\\
\bottomrule
\end{tabular}
\end{table}

\subsection{The long-horizon filtering floor}
\label{app:add-floor}

\begin{figure}[!htbp]
\centering
\includegraphics[width=0.85\textwidth]{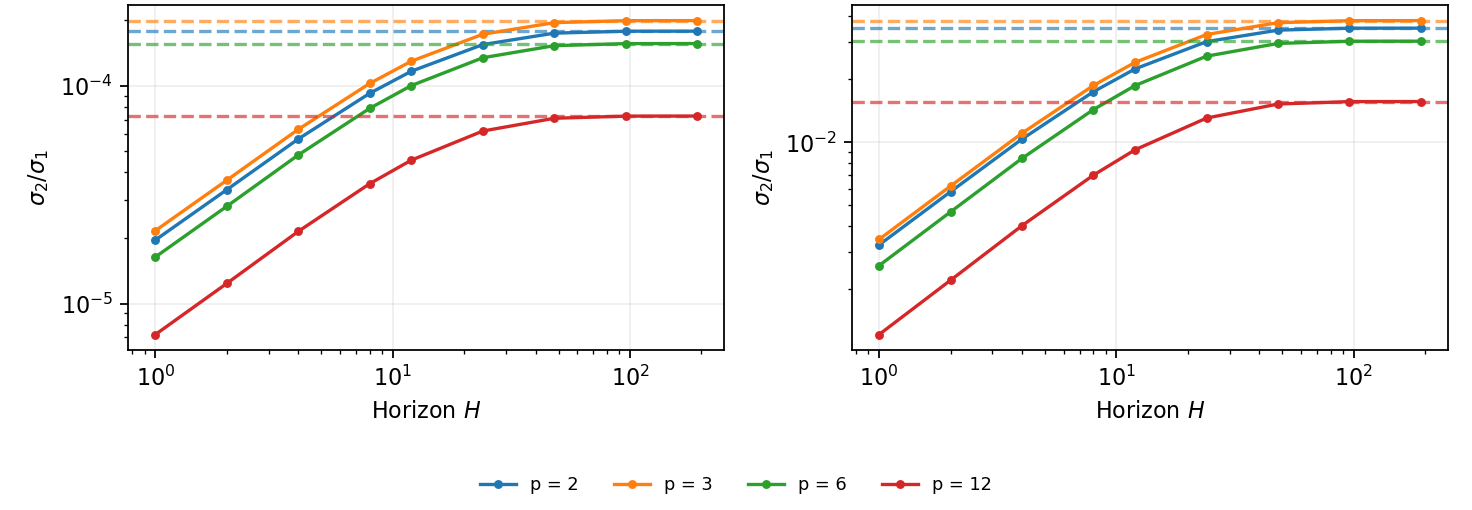}
\caption{Population ratio $\sigma_2/\sigma_1$ of the truncated infinite-past coefficient matrix $B_{H,p}$ against the horizon, $L=144$, logarithmic axes. Left: near-aligned design. Right: dense design. Dashed lines: the analytic long-horizon limits $\kappa_{\infty,1}(p)$ of Theorem~\ref{thm:floor}.}
\label{fig:add-floor}
\end{figure}

Figure~\ref{fig:add-floor} reports the population diagnostic \eqref{eq:simulation-relative-floor} for both observation designs at $L=144$, patch lengths $p\in\{2,3,6,12\}$, a rank-one budget, and horizons up to $H=192$. Nothing is trained. For every patch length the ratio $\sigma_2(B_{H,p})/\sigma_1(B_{H,p})$ increases with the horizon and converges to the positive limit $\kappa_{\infty,1}(p)$ of Theorem~\ref{thm:floor}, essentially reached by $H=96$. The operator-norm bound \eqref{eq:oprate} holds in all $72$ (design, patch, horizon) cases. The floor is modest in these systems, at most $0.038$ for the dense design and $2.0\cdot10^{-4}$ for the near-aligned design, while the predictable signal decays from $0.83$ at $H=1$ to about $2\cdot10^{-9}$ at $H=192$. This illustrates the distinction in \eqref{eq:absolute}: a positive relative floor accompanies a vanishing absolute approximation error.

\subsection{Head count when the target is nearly rank one}
\label{app:add-heads}

We train $m\in\{1,2,3,4\}$ heads on the dense design of Section~\ref{sec:experiments} with $L\in\{36,64,144,256,576,1024\}$ and $H\in\{1,2,4,8,12,16\}$ ($36$ settings), using the protocol of Appendix~\ref{app:add-protocol}. Here the finite-window target is close to rank one: at $p=\sqrt L$, an upper bound on the one-head approximation error is at most $12\%$ of the trained one-head excess risk in every setting.

\begin{figure}[!htbp]
\centering
\includegraphics[width=0.9\textwidth]{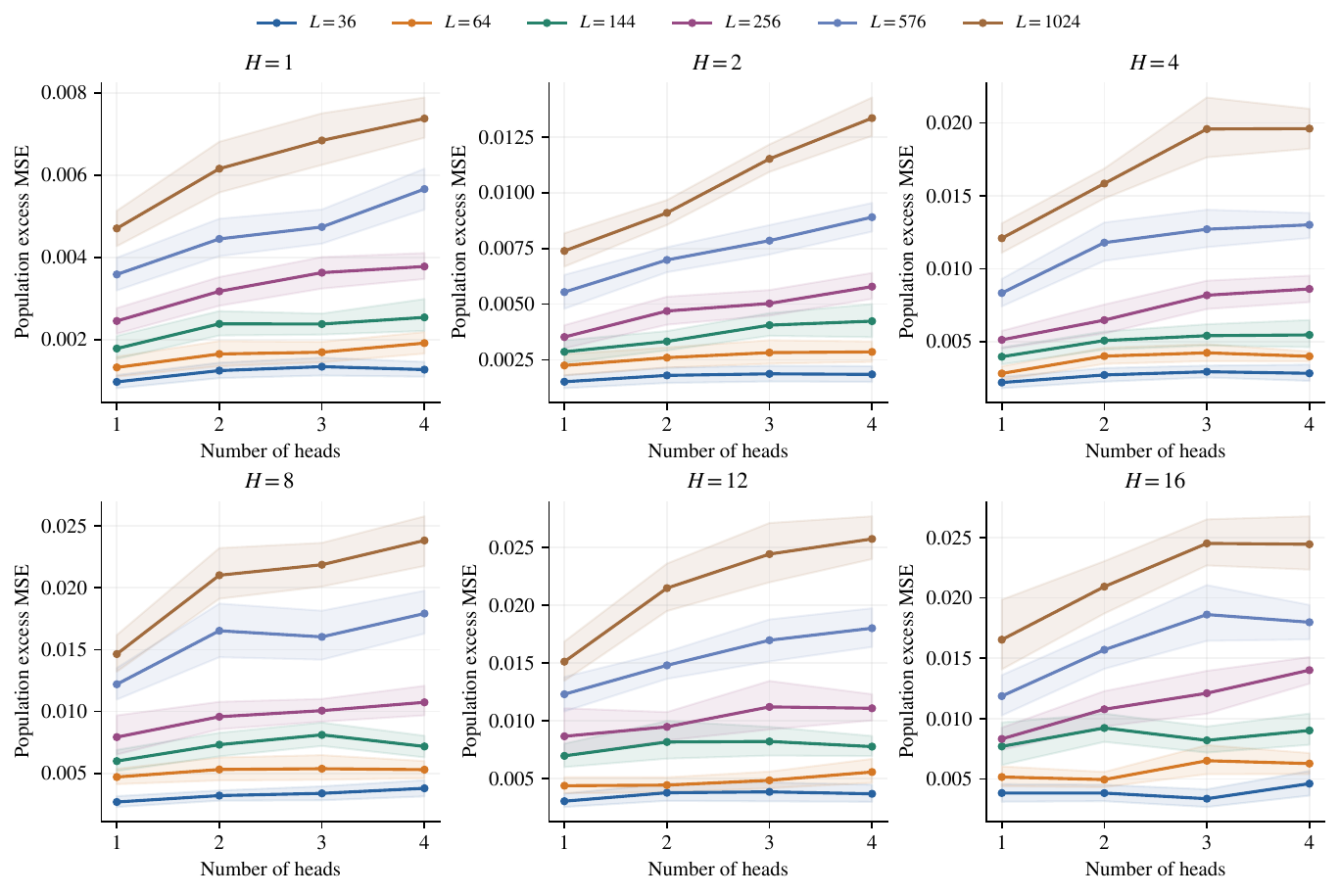}
\caption{Nearly rank-one target (dense design): population excess risk against the number of heads, with the patch selected by validation in each repetition. Panels show horizons, colors show window lengths, and shading shows pointwise $95\%$ bootstrap intervals over $20$ repetitions.}
\label{fig:add-heads}
\end{figure}

Additional heads enlarge the coefficient class, and its dimension $d_m(p)$ in \eqref{eq:dimension}, without a meaningful representation gain, so the estimation term dominates. With the patch chosen by validation, adding heads raises the mean excess risk relative to one head in almost every setting ($34$, $35$, and $36$ of the $36$ settings for two, three, and four heads). The paired $95\%$ interval for this change lies entirely above zero in most settings ($28$, $32$, and $30$ of $36$) and never lies entirely below zero (Figure~\ref{fig:add-heads}). The patch with the lowest mean risk stays close to $\sqrt L$ for every head count: $p=\sqrt L$ is within $5\%$ of the best mean risk in $115$ of the $144$ (head count, $L$, $H$) settings.

\subsection{Head count when the target has rank three}
\label{app:add-rank3}

To create a target that one head cannot represent, we use the delayed autoregression
\begin{equation}
X_{t+1}=c_1X_{t-4}+c_2X_{t-29}+c_3X_{t-58}+\varepsilon_{t+1},
\label{eq:add-rank3}
\end{equation}
written as a state-space model with the last $59$ observations as the state and no measurement noise. The latter is a boundary case of the assumption $R>0$, chosen to test the head-count restriction rather than the filtering results. The innovation variance is set so that $\operatorname{Var}(X_t)=1$. For $H\le4$ and $L\ge64$, the finite-window optimal forecast places weights $c_1,c_2,c_3$ on the lags $5-H$, $30-H$, and $59-H$. At $p=\sqrt L$ these three entries fall in distinct rows and columns of $B^\circ_{H,L,p}$, which therefore has rank three with singular values $(c_1,c_2,c_3)$. Since $\mathcal E_m^\Gamma(B)\ge\lambda_{\min}(\Gamma_{L,p})\sum_{k>m}\sigma_k(B)^2$, one and two heads face a positive approximation floor that three heads remove. We consider equal weights $(0.3,0.3,0.3)$ and unequal weights $(0.45,0.30,0.15)$, with $m\in\{1,2,3,4\}$, $L\in\{64,144,256,576\}$, and horizons up to $H=16$ (equal) or $H=12$ (unequal).

\begin{figure}[!htbp]
\centering
\includegraphics[width=0.8\textwidth]{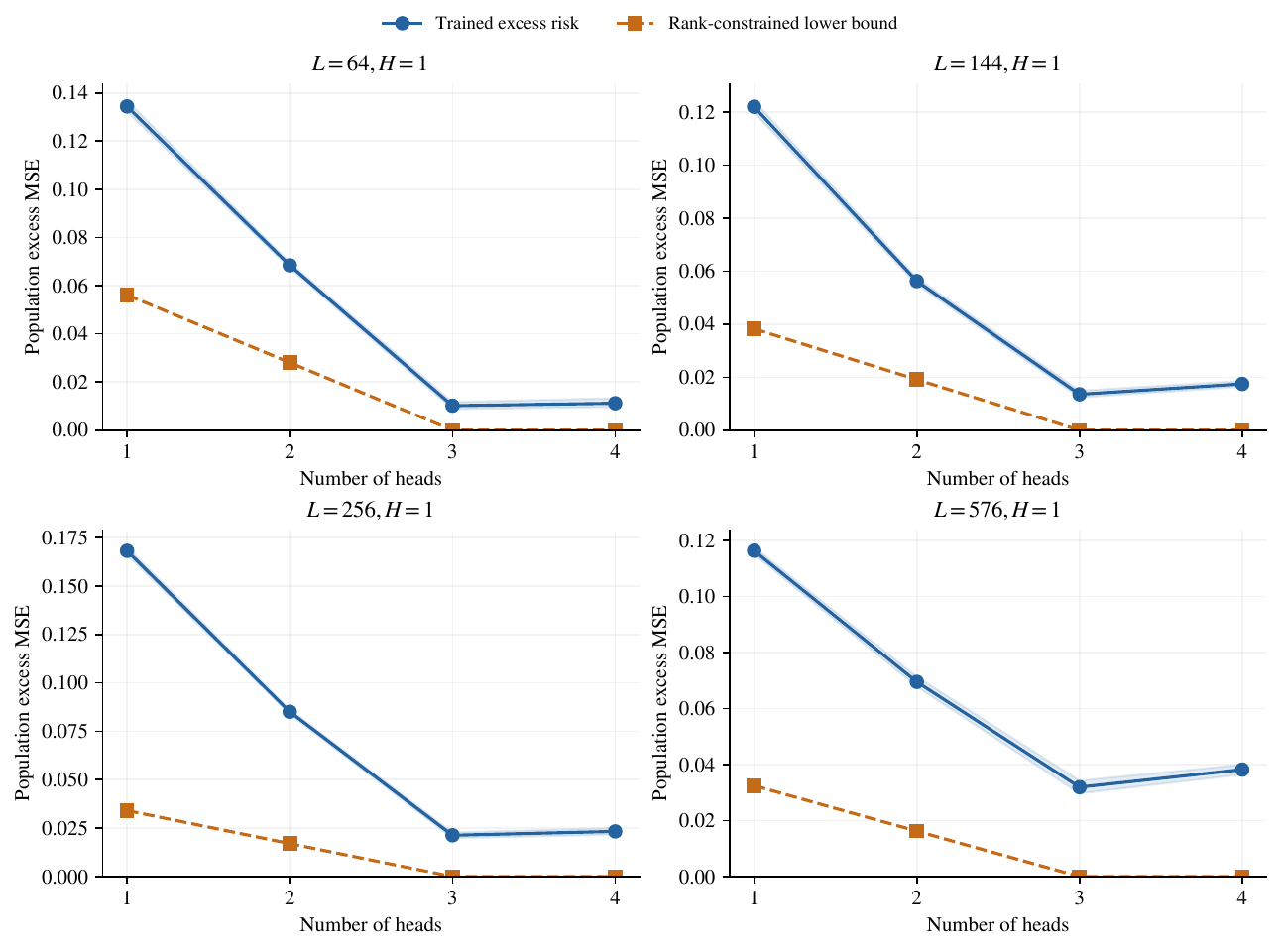}
\caption{Rank-three target with equal weights, $H=1$, $p=\sqrt L$: population excess risk of the trained model (blue) and the rank lower bound $\lambda_{\min}(\Gamma_{L,p})\,0.09\,(3-m)_+$ (orange) against the number of heads $m$, for four window lengths.}
\label{fig:add-rank3}
\end{figure}

Table~\ref{tab:add-rank3} and Figure~\ref{fig:add-rank3} show the approximation--estimation tradeoff in the head count. Up to the target rank, each additional head lowers the excess risk. At $L=64$, $H=1$, and $p=8$, the mean excess risk with equal weights is $0.134$, $0.068$, $0.010$, and $0.011$ for one to four heads. A fourth head, which adds no representation, never lowers the mean risk and raises it significantly in $10$ of $12$ settings. With unequal weights the third singular value is small, and the third head helps in $9$ of $12$ settings but is significantly worse than two heads in one ($L=576$, $H=4$).

\begin{table}[!htbp]
\centering
\caption{Rank-three targets: effect of adding one head at $p=\sqrt L$, over $12$ settings ($4$ windows, $H\in\{1,2,4\}$) and $20$ repetitions. ``Lower mean'' counts settings where the larger model has lower mean population excess risk. The interval columns count paired $95\%$ intervals entirely favoring the larger or the smaller model.}
\label{tab:add-rank3}
\begin{tabular}{@{}llccc@{}}
\toprule
Weights & Heads & Lower mean & Favors more heads & Favors fewer heads\\
\midrule
$(0.30,0.30,0.30)$ & 1 to 2 & 12/12 & 12/12 & 0/12\\
 & 2 to 3 & 12/12 & 12/12 & 0/12\\
 & 3 to 4 & 0/12 & 0/12 & 10/12\\
\midrule
$(0.45,0.30,0.15)$ & 1 to 2 & 12/12 & 12/12 & 0/12\\
 & 2 to 3 & 9/12 & 9/12 & 1/12\\
 & 3 to 4 & 0/12 & 0/12 & 10/12\\
\bottomrule
\end{tabular}
\end{table}

\subsection{Patch length across window lengths}
\label{app:add-scaling}

Section~\ref{sec:experiments} uses two window lengths. To examine how the preferred patch length grows with $L$, we train one head at nine window lengths $L\in\{36,64,144,256,576,784,1024,1296,1600\}$, all perfect squares, and six horizons $H\in\{1,2,4,8,12,16\}$, for four designs. These are the dense design of Section~\ref{sec:experiments} (baseline), $A=\operatorname{diag}(0.30,-0.60,0.98)$ with the same $C$ (changed $A$), $C=(0.1,2.0,0.1)$ with the same $A$ (changed $C$), and both changes together. For each design, window, and horizon, $\widehat p^\star$ minimizes the mean population excess risk over all divisors of $L$, and we fit $p=C\sqrt L$ through the origin, $\widehat C=\sum_L\sqrt L\,\widehat p^\star(L)/\sum_L L$, with intervals from $2{,}000$ paired bootstrap resamples that reselect $\widehat p^\star$ in each draw.

\begin{figure}[!htbp]
\centering
\includegraphics[width=\textwidth]{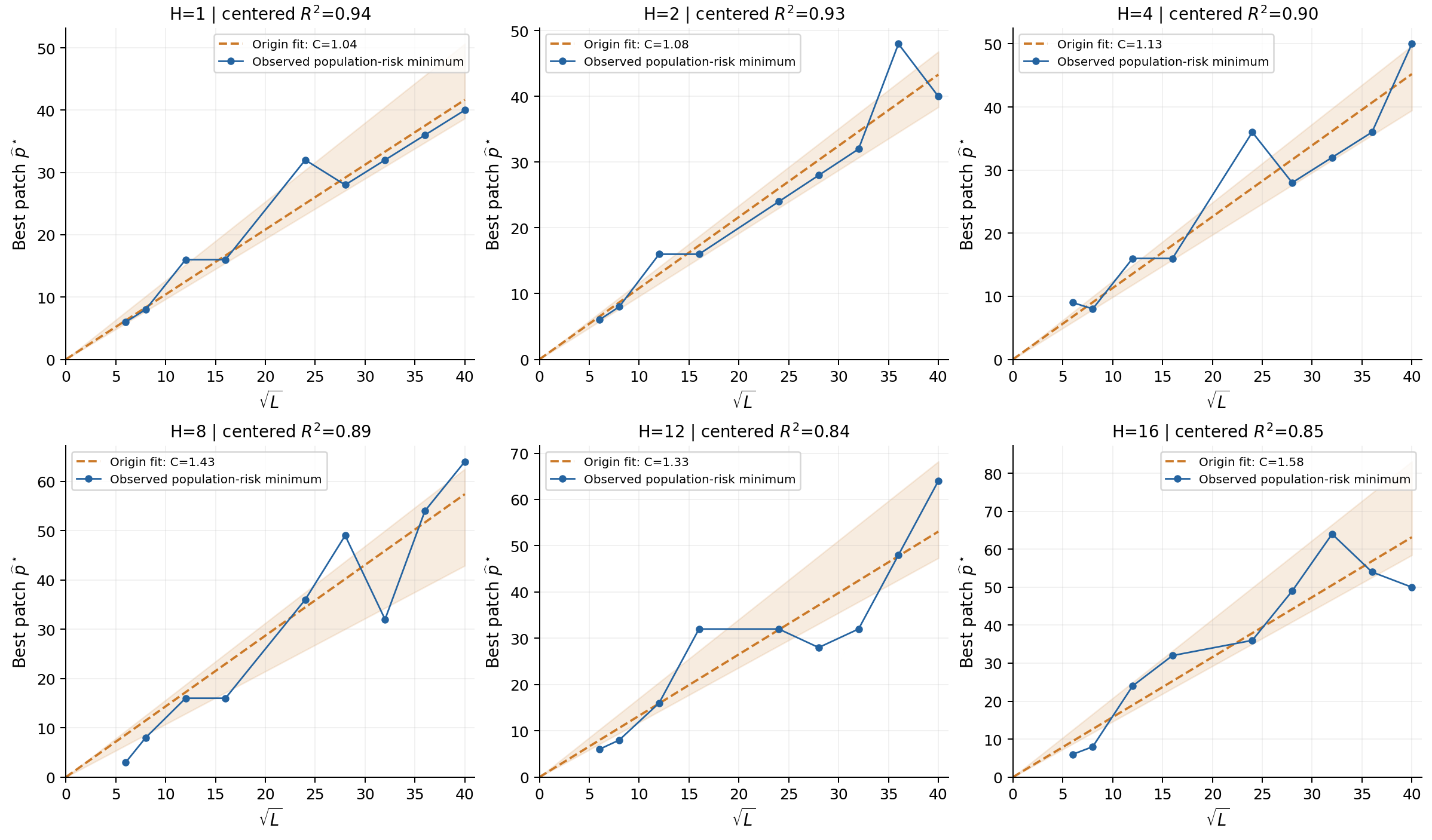}
\caption{Baseline (dense) design: the patch length $\widehat p^\star$ with the lowest mean population excess risk (blue) against $\sqrt L$ for nine window lengths, with the through-origin fit $p=\widehat C\sqrt L$ and its $95\%$ bootstrap band (orange). Panels: horizons, with centered $R^2$.}
\label{fig:add-window}
\end{figure}

Across the $216$ (design, $L$, $H$) conditions, $\widehat p^\star=\sqrt L$ in $98$, $p=\sqrt L$ is within $5\%$ of the best mean risk in $137$, and an extreme patch length is best only once. For the baseline, $\widehat p^\star$ tracks $\sqrt L$ closely (Figure~\ref{fig:add-window}): $\widehat C$ rises from $1.04$ at $H=1$ to $1.58$ at $H=16$, with centered $R^2\ge0.84$ (Table~\ref{tab:add-scaling}). The changed-$A$ and changed-$C$ designs behave similarly, except for a poor fit for changed $C$ at $H=16$. With both changes, the square-root relation fails at $H\ge8$: the preferred patches are much shorter ($\widehat C\approx0.13$) and the fit is worse than a constant. The square-root scale is therefore a useful reference whose constant depends on the system and the horizon, not a universal law, in line with the conditional statements of Section~\ref{sec:selection}.

\begin{table}[!htbp]
\centering
\caption{Fitted constants $\widehat C$ in $p=C\sqrt L$ over nine window lengths, with centered $R^2$ in parentheses (one head, $100$ training series, $20$ repetitions). Negative $R^2$ means that the fit is worse than a constant.}
\label{tab:add-scaling}
\begin{tabular}{@{}rcccc@{}}
\toprule
$H$ & Baseline & Changed $A$ & Changed $C$ & Changed $A$ and $C$\\
\midrule
1 & 1.04 (0.94) & 0.96 (0.83) & 1.06 (0.94) & 1.33 (0.64)\\
2 & 1.08 (0.93) & 1.00 (0.99) & 1.15 (0.94) & 1.06 (0.87)\\
4 & 1.13 (0.90) & 1.14 (0.94) & 0.98 (0.96) & 1.36 (0.75)\\
8 & 1.43 (0.89) & 1.41 (0.76) & 1.32 (0.85) & 0.13 ($-$0.74)\\
12 & 1.33 (0.84) & 1.66 (0.94) & 1.55 (0.78) & 0.13 ($-$0.73)\\
16 & 1.58 (0.85) & 1.60 (0.92) & 1.41 (0.34) & 0.14 ($-$0.66)\\
\bottomrule
\end{tabular}
\end{table}

\FloatBarrier

\end{document}